%% file: main.tex
\documentclass[10pt]{article} % For LaTeX2e
\usepackage[accepted]{tmlr}
\let\classAND\AND
\let\AND\relax
\input{math_commands.tex}

\usepackage{hyperref}
\usepackage{url}

\usepackage{microtype}
\usepackage{graphicx}
\usepackage{subfigure}
\usepackage{soul} % for strikethrough
\usepackage{booktabs} % for professional tables
\usepackage{amssymb}% http://ctan.org/pkg/amssymb
\usepackage{pifont}% http://ctan.org/pkg/pifont

\usepackage{enumitem}
\usepackage{wrapfig}
\usepackage{caption}
\usepackage{mdframed} % Add this to your preamble
\usepackage{comment}

\newcommand{\cmark}{\ding{51}}%
\newcommand{\xmark}{\ding{55}}%

\usepackage{amsmath}
\usepackage{amssymb}
\usepackage{mathtools}
\usepackage{amsthm}
\usepackage[capitalize,noabbrev]{cleveref}

\usepackage{algorithm}
\usepackage{algorithmic}

\let\AND\classAND

\AtBeginEnvironment{algorithmic}{\let\AND\algoAND}

\floatname{algorithm}{Algorithm}

\theoremstyle{plain}

\theoremstyle{definition}

\theoremstyle{remark}

\usepackage[textsize=tiny]{todonotes}
\usepackage{scalerel} % for smaller subscript 
\usepackage{xspace}

\newcommand{\method}{MoReFree\xspace}

\newcommand{\projectwebsite}{https://yangzhao-666.github.io/morefree}

\title{Reset-free Reinforcement Learning with World Models}

\author{\name Zhao Yang \email z.yang@liacs.leidenuniv.nl \\
      \addr The Leiden Institute of Advanced Computer Science \\
      Leiden University
      \AND
      \name Thomas M. Moerland \\
      \addr The Leiden Institute of Advanced Computer Science \\
      Leiden University
      \AND
      \name Mike Preuss \\
      \addr The Leiden Institute of Advanced Computer Science \\
      Leiden University 
      \AND
      \name Aske Plaat \\
      \addr 
      The Leiden Institute of Advanced Computer Science \\
      Leiden University 
      \AND
      \name Edward S. Hu \\
      \addr 
      GRASP Lab \\
      University of Pennsylvania
      }

\def\month{02}  % Insert correct month for camera-ready version
\def\year{2025} % Insert correct year for camera-ready version
\def\openreview{\url{https://openreview.net/forum?id=ZdMIXltJzK}} % Insert correct link to OpenReview for camera-ready version

\begin{document}

\maketitle

\begin{abstract}
Reinforcement learning (RL) is an appealing paradigm for training intelligent agents, enabling policy acquisition from the agent's own autonomously acquired experience. However, the training process of RL is far from automatic, requiring extensive human effort to reset the agent and environments.
% Most methods that attempt to remove one assumption rely on the presence of another to compensate. For instance, most reset-free approaches lean heavily on human-supervised guidance, while most unsupervised RL methods presume episodic resets.
To tackle the challenging reset-free setting, we first demonstrate the superiority of model-based (MB) RL methods in such setting, showing that a straightforward adaptation of MBRL can outperform all the prior state-of-the-art methods while requiring less supervision. We then identify limitations inherent to this direct extension and propose a solution called model-based reset-free (MoReFree) agent, which further enhances the performance. \method adapts two key mechanisms, exploration and policy learning, to handle reset-free tasks by prioritizing task-relevant states. It exhibits superior data-efficiency across various reset-free tasks without access to environmental reward or demonstrations while significantly outperforming privileged baselines that require supervision. Our findings suggest model-based methods hold significant promise for reducing human effort in RL. Website: \url{\projectwebsite}
\end{abstract}

\section{Introduction}
\label{intro}
% paragraph on RL, doing away with the assumptions
Reinforcement learning presents an attractive framework for training capable agents. At first glance, RL training appears intuitive and autonomous - once a reward is defined, the agent learns from its own automatically gathered experience. However, in practice, RL training often assumes the access to environmental resets that can require significant human effort to setup, which poses a significant barrier for real world applications of RL like robotics. 

Most RL systems on real robots to date have employed various strategies to implement resets, all requiring a considerable amount of effort \citep{levine2016end,yahya2017collective,zhu2019dexterous,nagabandi2020deep}. In \citet{nagabandi2020deep}, which trains a dexterous hand to rotate balls, the practitioners had to (1) position a funnel underneath the hand to catch dropped balls, and (2) deploy a separate robot arm to pick up the dropped balls for resets, and (3) script the reset behavior. These illustrate that even for simple behaviors, proper implementation of reset mechanisms can result in significant human effort and time. 

Rather than depending on human-engineered reset mechanisms, the agent can operate within a reset-free training scheme, learning to reset itself \citep{eysenbach2017leave,sharma2021vaprl,sharma2022state,haldar2023teach} or train a policy capable of starting from diverse starting states \citep{zhu2020ingredients}. However, the absence of resets introduces unique exploration challenges. Without periodic resets, the agent can squander significant time in task-irrelevant regions that require careful movements to escape and may overexplore, never returning from indefinite exploration. Recent unsupervised model-based RL (MBRL) approaches \citep{lexa2021, hu2023planning} in the episodic setting have shown sophisticated exploration, high data-efficiency and promising results in long-horizon tasks. This prompts the question: \textit{would MBRL agents excel in the reset-free RL setting?}

As an initial attempt, we first evaluate an unsupervised MBRL agent, in a reset-free Ant locomotion task. The ant is reset to the center of a rectangular arena, and is tasked with navigating to the upper right corner. The agent is reset only once at the start of training. The evaluation is episodic - the agent is reset at the start of each evaluation episode.

\begin{wrapfigure}[17]{}{0.45\textwidth}
\vspace{-0.5cm}
\hspace{-0.5cm}
 \centering
    \includegraphics[scale=0.05]{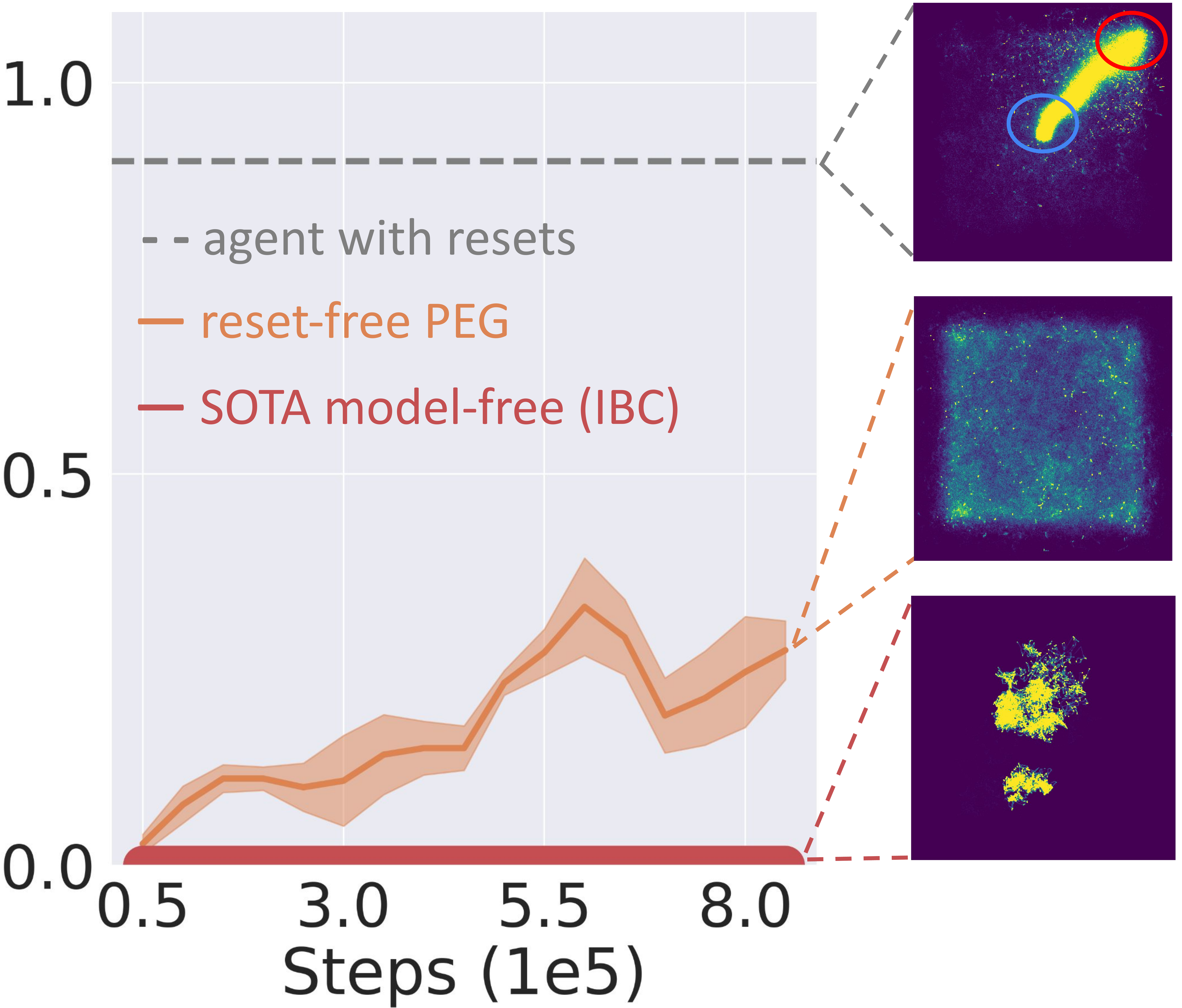}
    \caption{Performance and collected data of different agents on the reset-free Ant locomotion task.}
    \label{fig:peg}
\end{wrapfigure}

For the MBRL agent, we use PEG \citep{hu2023planning}, which was developed to solve hard exploration tasks in the episodic setting. As seen in \cref{fig:peg}, PEG, with minor modifications for the reset-free version, outperforms prior state-of-the-art, model-free agent, IBC \citep{kim2023free}, tailored for the reset-free setting. 

In \cref{fig:peg}, we plot state visitation heatmaps of the agents, where lighter colors correspond to more visitations. The oracle agent, with access to resets, explores the the ``task-relevant'' area between the initial and top right corner,  which is ideal for training a policy that succeeds in episodic evaluation. IBC's heatmap (bottom) shows that it fails to explore effectively, never encountering the goal states in the top right region. In contrast, PEG exhaustively explores the entire space, as seen through its uniform heatmap.  This results in  an overexploration problem - PEG may devote considerable time on finding irrelevant states rather than concentrating on the task-relevant region of the task. This leads us to ask: \textbf{how can MBRL agents acquire more task-relevant data in the reset-free setting to improve its performance?}

\begin{figure}[!htb]
    \centering
    \includegraphics[width=0.9\textwidth]{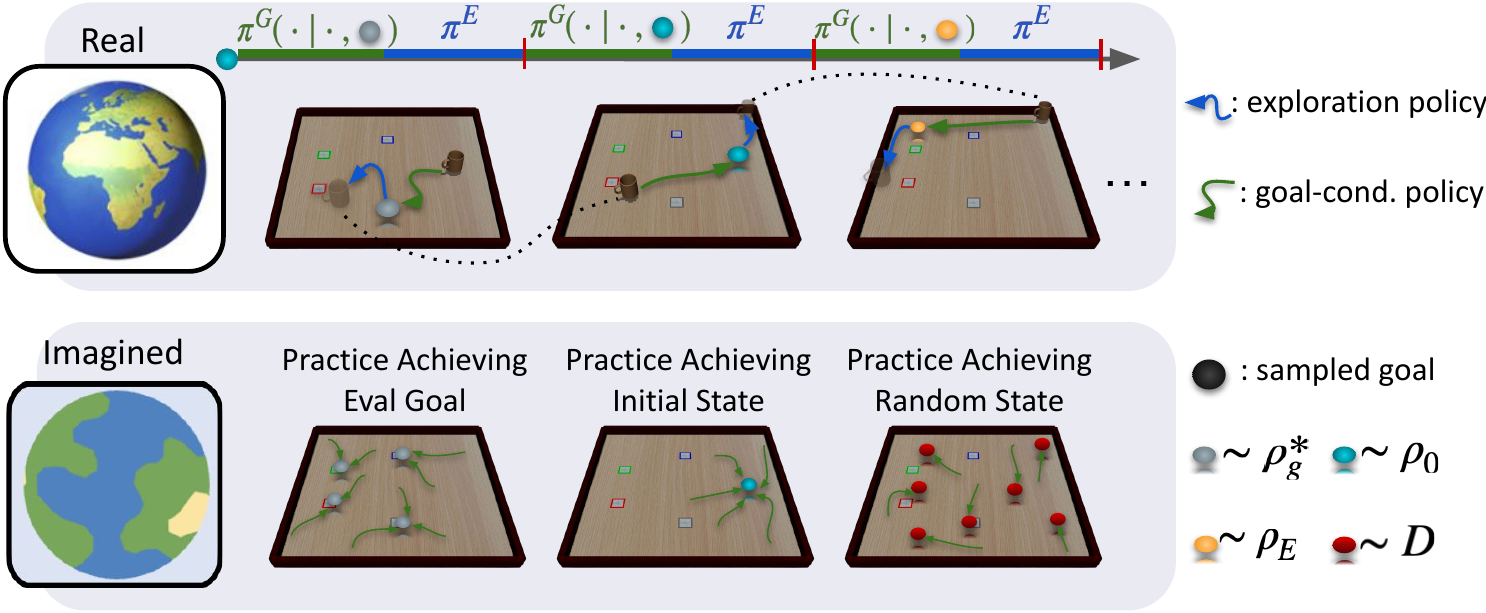}
    \caption{\method is a model-based RL agent for solving reset-free tasks. \textbf{Top row:} \method strikes a balance between exploring unseen states and practicing optimal behavior in task-relevant regions by directing the goal-conditioned policy to achieve evaluation states, initial state states (emulating a reset), and exploratory goals.
    \textbf{Bottom row:} \method focuses the goal-conditioned policy training inside the world model on achieving evaluation states, initial states, and random replay buffer states to better prepare the policy for the aforementioned exploration scheme.
    }
    \label{fig:concept}
\end{figure}

We propose \textbf{Mo}del-based, \textbf{Re}set-\textbf{Free} (\method), which improves two key mechanisms in model-based RL, exploration and policy optimization, to better handle reset-free training. Following the top row of \cref{fig:concept}: to gather task-relevant data without resets, we define a training curriculum that alternates between temporally extended phases of task solving, resetting, and exploration. Next, as seen in the bottom row of \cref{fig:concept}, we bias the policy training within the world model towards achieving task-relevant goals such as reaching initial states and evaluation states. 

Our key contributions are as follows: 
\textbf{(1)} We demonstrate the viability of using model-based agents with strong exploration abilities for the reset-free setting as well as their inherent limitations. We address such limitations through the \method framework which focuses exploration and policy optimization on task-relevant states. \textbf{(2)}  We evaluate the adapted reset-free version of MBRL baseline and \method against state-of-the-art reset-free methods in 8 challenging reset-free tasks ranging from manipulation to locomotion. 
Notably, both model-based approaches outperform prior state-of-the-art baselines in 7/8 tasks in final performance and data efficiency, all the while requiring less supervision (e.g. environmental reward or demonstrations).
\method outperforms the model-based baseline in the 3 hardest tasks.
\textbf{(3)}
We perform in-depth analysis of the \method and baselines behaviors, and show that \method  explores the state space thoroughly while retaining high visitation counts in the task-relevant regions. 
Our ablations show that the performance gains of \method come from the proposed design choices and justify the approach.

% our key contributions: 
% Our key contributions are: \textbf{(1)} We demonstrate superiority and limitations of model-based methods in the reset-free RL setting and propose \method, an enhanced model-based framework. An additional advantage of using MBRL is that, unlike model-free baselines, it does not require other forms of supervision (e.g. ground-truth reward functions or demonstrations), further reducing human effort and increasing autonomy of RL. \textbf{(2)} We compare the direct extension of MBRL and \method against baselines on eight challenging non-episodic, continuous control tasks ranging from pick-and-place manipulation to multi-legged ant locomotion. Two MBRL methods notably outperform model-free baselines with privileged access to more supervision in final performance and data-efficiency in 7/8 tasks, and \method beats the MBRL backbone in four more difficult tasks. This highlights the superiority of model-based approaches in the reset-free setting and effectiveness of the proposed solution. \textbf{(3)} We conduct extensive analysis of MBRL methods and baselines' exploration behavior. We find that \method's exploration strategy explores the state space thoroughly while retaining high visitation counts around task-relevant states. Our ablations show that the performance gains of \method are coming from the proposed design choices and therefore justifying the overall approach.

\section{Related Work}
\label{section:related work}

% In real-world applications like robotics, resets require extensive human effort, planning and cost towards implementation. 
% Some notable examples include creating motorized valves to enable automatic positional resets for a valve-turning task \citep{zhu2019dexterous}, or deploying a separate robot with a scripted primitive to reset fallen balls for baoding ball manipulation \citep{nagabandi2020deep}.
% \zyc{MEDAL, ibc also mentioned reward-free learning.}

\textbf{Reset-free RL:} There is a growing interest in researching reinforcement learning methods that can effectively address the complexities of reset-free training.
\cite{sharma2021autonomous} proposes a reset-free RL benchmark (EARL) and finds that standard RL methods like SAC \citep{haarnoja2018soft} fail catastrophically in EARL.
Multiple approaches have been proposed to address reset-free training, which we now summarize. One approach is to add an additional reset policy, to bring the agent back to suitable states for learning \citep{eysenbach2017leave, kim2022automating, sharma2021vaprl, sharma2022state, kim2023free}. LNT \citep{eysenbach2017leave} and \cite{kim2022automating} train a reset policy to bring the agent back to initial state distribution, supervised by dense rewards and demonstrations respectively.
MEDAL \citep{sharma2022state, sharma2023self}, train a goal-conditioned reset policy and direct it to reset  goal states from demonstrations.
IBC \citep{kim2023free} defines a curriculum for both task and reset policies without requiring demonstrations. 
VaPRL \citep{sharma2021vaprl} trains a single goal-conditioned policy to reach high value states close to the initial states.
% Increase the variance of the initial state distribution. Reset-free GPS, R3L
Instead of guiding the agent back to familiar states, R3L \citep{zhu2020ingredients} and \cite{xu2020continual} learn to reset the policy to diverse initial states, resulting in a policy that is more robust to variations in starting states. However, such methods are limited to tasks where exploration is unchallenging.
The vast majority of reset-free approaches are model-free, with a few exceptions \citep{lu2020adaptive,lu2020reset}.
Other works \citep{gupta2021reset, smith2019avid} model the reset-free RL training process as a multi-task RL problem and require careful definition of the task distribution such that the tasks reset each other. 

% Goal-conditioned Exploration:
% curriculum of goals , but not designed for episodic setting
%
\textbf{Goal-conditioned Exploration:} A common theme running through the aforementioned work is the instantiation of a curriculum, often through commanding goal-conditioned policies, to keep the agent in task-relevant portions of the environment while exploring. Closely related is the subfield of goal-conditioned exploration in RL, where a goal-conditioned agent selects its own goals during training time to generate data. There is a large variety of approaches for goal selection, such as task progress \citep{baranes2013active, veeriah2018many}, intermediate difficulty \citep{florensa2018automatic}, value disagreement \citep{zhang2020automatic}, state novelty \citep{pong2019skew, pitis2020maximum}, world model error \citep{hu2023planning, sekar2020planning}, and more. Many goal-conditioned exploration methods use the ``Go-Explore'' \citep{ecoffet2021first} strategy, which first selects a goal and runs the goal-conditioned policy (``Go''-phase), and then switches to an exploration policy for the latter half of the episode (``Explore''-phase). PEG~\citep{hu2023planning}, which \method uses, extends Go-Explore to the model-based setting, and utilizes the world model to plan states with higher exploration value as goals.
% Goal-conditioned exploration methods frequently outperform non-goal-directed counterparts, due to their structured exploration behavior. 
However, such methods are not designed for the reset-free RL setting, and may suffer from \textit{over-exploration} of task-irrelevant states.

\textbf{Learned Reward Functions: } Instead of requiring the environment to provide a reward function, the agent can learn its own reward function from onboard sensors and data. Given human specified example states, e.g. a goal image, VICE and C-Learning train reward classifiers over examples \citep{fu2018variational, eysenbach2021clearning} and agent data. The learned dynamical distance function \citep{hartikainen2019dynamical} learns to predict the number of actions between pairs of states. The dynamical distance function is used by unsupervised MBRL approaches like LEXA and PEG \citep{lexa2021, hu2023planning} to train the goal-conditioned policy. \method also employs the dynamical distance function as the reward function to eliminate the need of the environmental reward.

\begin{table}[!htb]
    \caption{
    A conceptual overview of reset-free methods. Existing methods are model-free, and most of them require other forms of supervision (environmental reward or demonstrations or both). In performance, \method improves over reset-free PEG, which significantly outperforms privileged baselines IBC, MEDAL and R3L.
    % R3L has relatively poor sample efficiency due to exploration and model-free learning and is outclassed by more privileged approaches like MEDAL and IBC. 
    % Most reset-free RL approaches assume access to rewards or demonstrations. R3L does not assume access to reward / demos, but has lackluster performance compared to MEDAL and VaPRL (which do assume access). In contrast, \method is a model-based approach that does not assume access and significantly outperforms the strongest privileged baselines MEDAL and IBC.
    }
    \centering
    %\scalebox{0.75}
    \begin{tabular}{ccccc|cc}
        Approach & MEDAL & IBC & VaPRL & R3L & reset-free PEG & MoReFree \\
        \hline
        Model-based  & \xmark & \xmark & \xmark & \xmark & \cmark & \cmark\\
        Demonstrations  & \cmark & \xmark & \cmark & \xmark & \xmark & \xmark \\
        Environmental reward  & \cmark & \cmark & \cmark & \xmark & \xmark & \xmark \\
    \end{tabular}
    
    \label{tab:assumptions}
\end{table}
% \tpc{Should use an em dash (—) for "only R3L matches—neither requires"}

We notice that the majority of all prior work are model-free and may suffer from poor sample efficiency and exploration issues. In contrast, our model-based approaches use world models to efficiently train policies and perform non-trivial goal-conditioned exploration with minimal supervision. See \cref{tab:assumptions} for a conceptual comparison between prior work and two model-based methods (\method and reset-free PEG).

\section{Preliminaries}
\subsection{Reset-free RL}

We follow the definition of reset-free RL from EARL~\citep{sharma2021autonomous}, and extend it to the goal-conditioned RL setting. Consider the goal-conditioned Markov decision process (MDP) $\mathcal{M}=(\mathcal{S}, \mathcal{G}, \mathcal{A}, p, r, \rho_0, \rho_{g^*}, \gamma)$. At each time step $t$ in the state $s_{t}\in \mathcal{S}$, a goal-conditioned policy $\pi(\cdot|s_{t}, g)$ under the goal command $g\in \mathcal{G}$ selects an action $a_t \in \mathcal{A}$ and transitions to the next state $s_{t+1}$ with the probability $p(s_{t+1}|s_t, a_t)$, and gets a reward $r(s_t, a_t, g)$.  $\rho_0$ is the initial state distribution, $\rho_{g^*}$ is the evaluation goal distribution, and $\gamma$ is the discount factor. 

The learning algorithm $\mathbb{A}$ is defined: $\{{s_i,a_i,s_{i+1}}\}_{i=0}^{t-1}\mapsto (a_t, \pi_t)$, which maps the transitions collected until the time step $t$ to the action $a_t$ the agent should take in the non-episodic training and the best guess $\pi_t$ of the optimal policy $\pi^*$ on the evaluation goal distribution ($\rho_{g^*}$). In reset-free training the agent will only be reset to the initial state $s_0 \sim \rho_0$ a few times. The evaluation of agents is still episodic. The agent always starts from $s_0 \sim \rho_0$, and is asked to achieve $g \sim \rho_{g^*}$. The evaluation objective for a policy $\pi$ is:
% \zyc{are these subscripts too small?}
\begin{equation}
 J(\pi)=\E_{s_0\sim \rho_0, g\sim \rho_{g^*}, a_j \sim \pi(\cdot| s_j,g),s_{j+1}\sim p(\cdot|s_j,a_j)}[\sum_{j=0}^{T} \gamma^j r(s_j,a_j,g)],
\end{equation} where $T$ is the total time steps during the evaluation. The goal of algorithm $\mathbb{A}$ during the reset-free training is to minimize the performance difference $\mathbb{D(A)}$ of the current policy $\pi_t$ and the optimal policy $\pi^*$:
% \tm{This might be a bit confusing definition of regret, which is often the (cumulative) reward of an optimal policy minus the actual obtained one? Since $J(\pi)$ is defined above as the regret during evaluation episodes, I'm not sure whether you still need the sum over $t$ below? Or the other way around, if you keep the sum over timesteps below, wouldn't you then just need the difference in expected reward between $\pi^\star$ and $\pi_t$?}
\begin{equation}
    \mathbb{D(A)}=\sum_{t=0}^\infty (J(\pi^*)-J(\pi_t)).
\end{equation}
In summary, the algorithm $\mathbb{A}$ should output an action $a_t$ that the agent should take in the non-episodic data collection and a policy $\pi_t$ that can maximize $J(\pi_t)$ at every time step $t$ based on all previously collected data.

\subsection{Model-based RL setup}
\label{section:mbge}
% \ehc{First motivate why we pick PEG / LEXA as the backbone RL agent. Because it already is unsupervised GCRL and can explore very effectively.}
Recent goal-conditioned MBRL approaches like LEXA \citep{lexa2021} and PEG \citep{hu2023planning} train goal-conditioned policies purely using synthetic data generated by learned world models. Their robust exploration demonstrates significant success in solving long-horizon goal-conditioned tasks. In the reset-free setting, strong exploration is crucial, as the agent can no longer depend on episodic resets to bring it back to task-relevant areas if it gets stuck.  Therefore, we select PEG as the backbone MBRL agent for its strong exploration abilities and sample efficiency.

PEG~\citep{hu2023planning} is a model-based Go-Explore framework that extends LEXA \citep{lexa2021}, an unsupervised goal-conditioned variant of DreamerV2 \citep{hafner2020dreamerv2}. 
The following components are parameterized by $\theta$ and learned:
\begin{equation}
\begin{split}
        &\text{world model: } \widehat{\mathcal{T}}_\theta(s_t|s_{t-1}, a_{t-1}) \\
        &\text{goal conditioned policy: } \pi^G_\theta(a_t|s_t,g) \hspace{0.9cm} \text{goal conditioned value: } V^G_\theta(s_t,g) \\
        &\text{exploration policy: } \pi^E_\theta(a_t|s_t) \hspace{2cm} \text{exploration value: } V^E_\theta(s_t) \\
    \end{split}
\end{equation}

The world model is a recurrent state-space model (RSSM) which is trained to predict future states and is used as a learned simulator to train the policies and value functions. 
The goal-conditioned policy $\pi^G_\theta$ is trained to reach random states sampled from the replay buffer. The exploration policy $\pi^E_\theta$ is trained on an intrinsic motivation reward that rewards world model error, expressed through the variance of an ensemble \citep{sekar2020planning}(see \cref{supp:p2e} for more details). Both policies are trained on simulated trajectory rollouts in the world model.

$\blacktriangleright$ \textbf{Self-supervised Goal-reaching Reward Function:} 
% The goal-conditioned policy $\pi^G$ is trained to reach random goal states sampled from the replay buffer, using a 
Rather than assuming access to the environmental reward, PEG learns its own reward function. PEG uses a dynamical distance function \citep{hartikainen2019dynamical} as the reward function within world models, which predicts the number of actions between a start and goal state. The distance function is trained on random state pairs from imaginary rollouts of $\pi^G_\theta$. $\pi^G_\theta$ is then trained to minimize the dynamical distance between its states and commanded goal state in imagination. See \cref{supp:reward fn} for more details.

%  The goal-conditioned reward is to a learned dynamical distance function.
% learned temporal reward function~\citep{hartikainen2019dynamical} that predicts the number of steps needed from the current state to the goal state. Crucially, this goal-conditioned reward function is self-supervised, and does not require an environmental reward function. 

\begin{wrapfigure}[15]{r}{0.25\textwidth} % Adjust the width as needed
\vspace{-0.8cm}
\hspace{-0.5cm}
    \centering
    \begin{minipage}{0.25\textwidth} % Fine-tune width here
        \begin{algorithm}[H]
            \caption{Go-Explore}
            \label{alg:mbge}
            \begin{algorithmic}[1]
                \STATE \textbf{Input:} $g,\pi^G_{\theta},\pi^E_{\theta}$
                \STATE $\tau_g \leftarrow \{\};\tau_e \leftarrow \{\}$
                \FOR{$t=1$ to $H_G$}
                    \STATE $a_t \sim \pi^G_\theta(\cdot~|s_t,g)$
                    \STATE $s_{t+1} \sim \mathcal{T}(\cdot~|s_t,a_t)$
                    \STATE $\tau_g \leftarrow \tau_g \cup\{s_t\}$
                \ENDFOR
                \FOR{$t=1$ to $H_E$}
                    \STATE $a_t \sim \pi^E_\theta(\cdot~|s_t)$
                    \STATE $s_{t+1} \sim \mathcal{T}(\cdot~|s_t,a_t)$
                    \STATE $\tau_e \leftarrow \tau_e \cup\{s_t\}$
                \ENDFOR
                \STATE \textbf{return} $\tau_g, \tau_e$
            \end{algorithmic}
        \end{algorithm}
    \end{minipage}
\end{wrapfigure}

$\blacktriangleright$ \textbf{Phased Exploration via Go-Explore:}
For data-collection, PEG employs the Go-Explore strategy.
In the ``Go''-phase, a goal is sampled from some goal distribution $\rho$. The goal-conditioned policy, conditioned on the goal is run for some time horizon $H_G$, resulting in trajectory $\tau_g$. 

Then, in the ``Explore''-phase, starting from the last state in the ``Go''-phase, the exploration policy is run for $H_E$ steps, resulting in $\tau_e$. The interleaving of goal-conditioned behavior with exploratory behavior results in more directed exploration and informative data. This in turn improves accuracy of the world model, and the policies that train inside the world model. 
See~\cref{alg:mbge} and~\cref{alg:peg} for pseudocode. The choice of goal distribution $\rho$ is important for Go-Explore. In easier tasks, the evaluation goal distribution $\rho_{g^*}$ may be sufficient. But in longer-horizon tasks, evaluation goals may be too hard to achieve. Instead, intermediate goals from an exploratory goal distribution $\rho_E$ can help the agent explore. We choose PEG, which generates goals by planning through the world model to maximize exploration value (see \cref{supp:peg} for details).

\begin{comment}
\setcounter{algorithm}{1}
\begin{algorithm}
        \caption{MBRL Backbone (PEG)}
        \label{alg:mbge}
        \begin{algorithmic}[1]
            \STATE \textbf{Input:} $\pi^G_\theta$, $\pi^E_\theta$, world model $\widehat{\mathcal{T}}_\theta$, goal distribution $\rho$
            \STATE \textbf{for} episode $i=1$ to $N$ \textbf{do}:
            \STATE \hspace{4pt} sample a goal $g~\sim \rho$
            \STATE \hspace{4pt} $\tau_g, \tau_e \leftarrow $Go-Explore($g, \pi^G, \pi^E$)   
            % \STATE \hspace{4pt} collect traj. $\tau^g$ using $\pi^G_\theta(\cdot \mid \cdot, g)$
            % \STATE \hspace{4pt} collect traj. $\tau^e$ using $\pi^E_\theta$
            \STATE \hspace{4pt} $\mathcal{D} \leftarrow \mathcal{D} \cup\tau^g \cup\tau^e$
            \STATE \hspace{4pt} update $\widehat{\mathcal{T}}_\theta$ with $\mathcal{D}$
            \STATE \hspace{4pt} update $\pi^G_\theta$ and $\pi^E_\theta$ with $\widehat{\mathcal{T}}_\theta$ in imagination
        \end{algorithmic}
\end{algorithm}
\end{comment}

\section{Method}

% \ehtext{Flow should be: 
% 1) We now introduce \method, a model-based approach to reset-free RL. We would like to utilize the benefits of model-based approaches, like improved goal-conditioned exploration through PEG, or sample-efficient gc-policy training in imagination from Dreamer / LEXA. 
% 2) But such approaches were designed for episodic RL. Simply porting them over may not work (and indeed in our ablations we find that to be the case.
% 3) Summarize \method. We identify two key parts of MBRL that need to be adapted. Exploration and Policy Training. 
% }

As motivated in \cref{intro} and \cref{fig:peg}, the direct application  of PEG to the reset-free setting shows promising performance but suffers from over-exploration of task-irrelevant states. To adapt model-based RL to the reset-free setting, we introduce \method, a model-based approach that improves PEG to handle the lack of resets and overcome the over-exploration problem. \method improves two key mechanisms of MBRL for reset-free training: exploration and policy training. 

\subsection{Back-and-Forth Go-Explore}
\label{section:exploration}

First, we introduce \method's procedure for collecting new datapoints in the real environment.
PEG \citep{hu2023planning} already has strong goal-conditioned exploration abilities, but was developed for solving episodic tasks. Without resets, PEG's Go-Explore procedure can undesirably linger in unfamiliar but task-irrelevant portions of the state space. This generates large amounts of uninformative trajectories, which in turn degrades world model learning and policy optimization. 

\begin{wrapfigure}[19]{r}{0.6\textwidth} % Adjust the width as needed (here 65% of text width)
\vspace{-0.8cm}
\hspace{-0.5cm}
    \centering
    \begin{minipage}{0.6\textwidth} % Fine-tune the inner width
        \begin{algorithm}[H]
            \caption{MBRL Backbone}
            \label{alg:peg}
            \begin{algorithmic}[1]
                \STATE \textbf{Input:} $\pi^G_\theta$, $\pi^E_\theta$, world model $\widehat{\mathcal{T}}_\theta$, goal distribution $\rho$ (including: exploratory goal distribution $\rho_E$, evaluation goal distribution $\rho_{g^*}$, initial state distribution $\rho_0$)
                \STATE $\mathcal{D} \leftarrow \{\}$
                \WHILE{within the reset-free horizon}
                    \STATE \hspace{4pt} \textcolor{orange}{\# reset-free PEG}
                    \STATE \hspace{4pt} \textcolor{orange}{sample a goal $g \sim \rho_E$}
                    \STATE \hspace{4pt} \textcolor{orange}{$\tau_g, \tau_e \leftarrow $Go-Explore($g, \pi^G_\theta, \pi^E_\theta$)}
                    \STATE \hspace{4pt} \textcolor{cyan}{\# MoReFree}
                    \STATE \hspace{4pt} \textcolor{cyan}{$\tau_g, \tau_e \leftarrow $Back-and-Forth Go-Explore($\pi^G_\theta, \pi^E_\theta, \rho)$}
                    \STATE \hspace{4pt} $\mathcal{D} \leftarrow \mathcal{D} \cup \tau_g \cup \tau_e$
                    \STATE \hspace{4pt} update $\widehat{\mathcal{T}}_\theta$ with $\mathcal{D}$
                    \STATE \hspace{4pt} update $\pi^E_\theta$ with $\widehat{\mathcal{T}}_\theta$ in imagination
                    \STATE \hspace{4pt} update $\pi^G_\theta$ with $\widehat{\mathcal{T}}_\theta$ in imagination, cond. on goals $g'$:
                    \STATE \hspace{4pt} \textcolor{orange}{$g' \sim \mathcal{D}$}
                    \STATE \hspace{4pt} \textcolor{cyan}{$g' \sim \mathcal{D}$ with $\Pr = 1-\alpha$, $g' \sim \rho_{g^*}, \rho_{0}$ with $\Pr = \alpha$}
                \ENDWHILE
            \end{algorithmic}
        \end{algorithm}
    \end{minipage}
\end{wrapfigure}

\method overcomes this by periodically directing the agent to return to the states relevant to the task (i.e. initial and evaluation goals).  We call this exploration procedure ``Back-and-Forth Go-Explore'', where we sample pairs of initial and evaluation goals and ask the agent to cycle back and forth between the goal pairs, periodically interspersed with exploration phases (see \cref{fig:concept} top row).

Now, we define the ``Back-and-Forth Go-Explore'' strategy as seen in~\cref{alg:bfge}. First, we decide whether to perform initial / evaluation state directed exploration. With probability $\alpha$, we sample goals $(g^*, g_0)$ from $\rho_{g^*}, \rho_0$ respectively. Then, we execute the Go-Explore routine for each goal. We name Go-Explore trajectories conditioned on initial state goals as ``Back'' trajectories, and Go-Explore trajectories conditioned on evaluation goals as ``Forward'' trajectories. With probability $1-\alpha$, we execute exploratory Go-Explore behavior by sampling exploratory goals from PEG. The difference between reset-free PEG and \method can be seen in Algorithm 2, unlike PEG, \method employs the ``Back-and-Forth Go-Explore''. 

By following this exploration strategy, the agent modulates between various Go-Explore strategies, alternating between solving the task by pursuing evaluation goals, resetting the task by pursuing initial states, and exploring unfamiliar regions via exploratory goals. 

% (Go-Explore($g_1, \pi^G, \pi^E$) followed by Go-Explore($g_2, \pi^G, \pi^E$)). 

% The agent will cycle between returning to the initial state distribution (``Back''), achieving evaluation goals (``Forth''), and exploring new states. This is achieved by setting the goal sampling distribution $\rho$ of Go-Explore (\cref{alg:mbge}) to sample between PEG exploratory goal distribution $\rho_E$, evaluation goal distribution $\rho_{g^*}$, and the initial state distribution $\rho_0$. 

% The agent explores the environment using Go-Explore, which switches between a goal-directed policy (``Go''-phase) and an exploration policy (``Explore''-phase). During the Go-phase, the choice of a goal distribution out of $\{ \rho_0, \rho_{g^*}, \rho_E\}$ is chosen with probabilities $\alpha, \alpha, 1-2\alpha$\tm{maybe also add the brackets here?}, and then a goal $g$ is sampled from the picked distribution.\ehc{Please check if alpha is correct.}\zyc{modified}
% \zytext{During the Go-phase, back-and-forth is chosen with probability $\alpha$, then a goal $g$ is in turn sampled from {$\rho_0, \rho_g$} to perform ``back'' and ``forth''. Exploratory goal distribution $\rho_E$ is chosen with probability $1-\alpha$.}
% After running the Go-phase with $g$ and the subsequent Explore-phase, we repeat the goal-sampling procedure for the next round of Go-Explore. See \cref{fig:concept}'s top row for a visualization of this continuous Go-Explore process with varying goals for each Go-phase.

\subsection{Learning to Achieve Relevant Goals in Imagination}

\begin{wrapfigure}[16]{r}{0.4\textwidth} % Change "l" to "r" for right side; adjust width as desired
\vspace{-0.8cm}
\hspace{-0.5cm}
    \centering
    \begin{minipage}{0.4\textwidth}
        \begin{algorithm}[H]
            \caption{Back-and-Forth Go-Explore}
            \label{alg:bfge}
            \begin{algorithmic}[1]
                \STATE \textbf{Input:} $\pi^G_\theta$, $\pi^E_\theta$, $\rho_{g^*}$, $\rho_0$, $\rho_{E}$
                \STATE Generate a random number $r$ in $[0, 1]$
                \IF{$r < \alpha$}
                    \STATE $g^*, g_0 \sim \rho_{g^*}, \rho_0$
                    \STATE $\tau_{g^*}, \tau^1_{e} \leftarrow$ Go-Explore($g^*, \pi^G_\theta, \pi^E_\theta$)
                    \STATE \# Continue from the terminal state of the previous Go-Explore.
                    \STATE $\tau_{g_0}, \tau^2_{e} \leftarrow$ Go-Explore($g_0, \pi^G_\theta, \pi^E_\theta$)
                    \STATE $\tau_g \leftarrow \tau_{g^*} \cup \tau_{g_0}$; $\tau_e \leftarrow \tau^1_{e} \cup \tau^2_{e}$
                \ELSE 
                    \STATE $g \sim \rho_{E}$
                    \STATE $\tau_{g}, \tau_{e} \leftarrow$ Go-Explore($g, \pi^G_\theta, \pi^E_\theta$)
                \ENDIF
                \STATE \textbf{return} $\tau_g, \tau_e$
            \end{algorithmic}
        \end{algorithm}
    \end{minipage}
\end{wrapfigure}

Next, we describe how \method  trains the goal-conditioned policy in the world model. To train $\pi^G_\theta$, \method samples various types of goals and executes $\pi^G_\theta(\cdot \mid \cdot, g)$ inside the world model to generate ``imaginary'' trajectories. The trajectory data is scored using the learned dynamical distance reward mentioned in \cref{section:mbge} 
% \tm{this was introduced before, and called learned dynamical distance function there. I would write "the learned temporal distance function introduced in Sec ..." here. } 
, and the policy is updated to maximize the expected return. This procedure is called imagination \citep{hafner2019dreamer}, and allows the policy to be trained on vast amounts of synthetic trajectories to improve sample efficiency. 

First, we choose to sample evaluation goals from $\rho_{g^*}$ since the policy will be evaluated on its evaluation goal-reaching performance. Next, recall that Back-and-Forth Go-Explore procedure also samples initial states from $\rho_0$ as goals for the Go-phase to emulate resetting behavior. Since we would like $\pi^G_\theta$ to succeed in such cases so that the task is reset, we will also sample from $\rho_0$.
Finally, we sample random states from the replay buffer to increase $\pi^G_\theta$'s ability to reach arbitrary states. The sampling probability for each goal type is set to $\alpha/2, \alpha/2, 1-\alpha$ respectively.
% See \cref{fig:concept} bottom row for an example.
%\zytext{The sampling probability for each goal type is set to $\alpha/2, \alpha/2, 1-\alpha$ respectively.}
In other words, \method biases the goal-conditioned policy optimization procedure to focus on achieving task-relevant goals (i.e. evaluation and initial states), as they are used during evaluation and goal-conditioned exploration to condition the goal-reaching policy (see \cref{fig:concept} bottom row). This leads to additional changes of line 13 in Algorithm 2.

\subsection{Implementation Details}
Our work builds on the top of PEG~\citep{hu2023planning}, and we use its default hyperparameters for world model, policies, value functions and temporal reward function. We set the length of each phase for Go-Explore $(H_G, H_E)$ to half the evaluation episode length for each task. We set the default value of $\alpha=0.2$ for all tasks (never tuned).
% we use the same setting for all tasks: $\alpha = 0.2$.\tm{You will vary $\alpha$ right, it's your most important hyperparameter? I would say it's 0.2 as a default value, unless otherwise specified.} 
See \cref{supp:hyperparameters} for more details and the supplemental for \method code.

% For MoReFree, we set the length of every chunk to $H_E$ equals to the episode length during the evaluation. Within each chunk, two policies are executed with half chunk ($H_E/2$ steps). At the beginning of each chunk, $\rho=\rho_0 \cup \rho_{g^*}$ with probability of $0.2$, otherwise $\rho=\rho_E$. For policy training, we sample $0.2$ of goals from $\rho_0 \cup \rho_{g^*}$, while the rest is sampled uniformly from the replay buffer $\mathcal{D}$. We keep these setting cross all tasks, unless otherwise noted. The codebase is included in the supplemental. 

\section{Experiments}
% \ehc{TODO: rewrite with DR3L perspective.}
We evaluate three MBRL methods (PEG \citep{hu2023planning}, the extension reset-free PEG and our proposed method \method) and four competitive reset-free baselines on eight reset-free tasks. We aim to address the following questions: 1) Do MBRL approaches work well in reset-free tasks in terms of sample efficiency and performance? 2) What limitations arise from running MBRL in the reset-free setting, and does our proposed solution \method address them? 3) What sorts of behavior do \method and baselines exhibit in such tasks, and are our design choices for \method justified?

\textbf{Baselines:} 
All baselines except for R3L are implemented using official codebases, see \cref{supp:baselines} for details.

% \ehc{Can you check if MEDAL (and others) description seems correct?}\zyc{checked}
\begin{itemize}
    \item 
    \textbf{PEG} \citep{hu2023planning} is the original episodic PEG in which exploratory goals are only sampled once at the beginning of each episode (in the reset-free setting, the episode is extremely long). The goal-conditioned policy and the exploration policy are then executed for the first half and second half of the episode, respectively. 
    \item 
    \textbf{reset-free PEG} is a straightforward extension of PEG to the reset-free setting. Exploratory goals are sampled every $H_G+H_E$ steps. Then, the goal-conditioned policy is executed for $H_G$ steps followed by the exploration policy being executed for $H_E$ steps.
    \item 
    \textbf{DreamerV2} \citep{hafner2020dreamerv2} is a commonly used MBRL method. The goal-conditioned policy is executed for the whole reset-free episode.
    \item \textbf{MEDAL} \citep{sharma2022state} requires demonstrations and trains two policies, one for returning to demonstration states and another that achieves task goals.
    \item \textbf{IBC} \citep{kim2023free} is a competitive baseline that outperforms prior reset-free work (e.g. MEDAL, VaPRL) by defining a bidirectional curriculum for the goal-conditioned forward and backwards (i.e. reset) policies trained using the environmental reward.
    \item \textbf{R3L} \citep{zhu2020ingredients} trains two policies, one for achieving task goals and another that perturbs the agent to novel states. Notably, it is the only baseline that operates without any additional assumptions (i.e. environmental rewards, demonstrations, and resets).
    \item \textbf{Oracle} is SAC \citep{haarnoja2018soft} trained under the episodic setting on the environmental reward.
\end{itemize}
Note that most baselines enjoy some advantage over two MBRL methods: MEDAL, IBC and Oracle use ground truth environmental reward, while MEDAL also uses demonstrations and Oracle uses resets.
See \cref{tab:assumptions} for a conceptual comparison between \method and prior work.

% \ehc{Mention R3L, while it is a baseline that fits the DR3L setting, it has notably poor performance compared to others and doesn't have a public implementation.}

% Since MoReFree is the first model-based reset-free method, baselines used for comparison are all model-free. \textbf{IBC}~\cite{kim2023free} is the current strongest method proposed for reset-free tasks and doesn't require any demonstrations. It learns to reach bidirectional curriculums towards the initial and goal states. 
% \textbf{MEDAL}~\cite{sharma2022state} leverages demonstrations and learns to get back to states in the demonstrations to start from. \textbf{Oracle} is a standard SAC~\cite{haarnoja2018soft} agent training under episodic setting which is used as a reference. 

\begin{figure}[!hbt]
    \centering
    \includegraphics[scale=0.4]{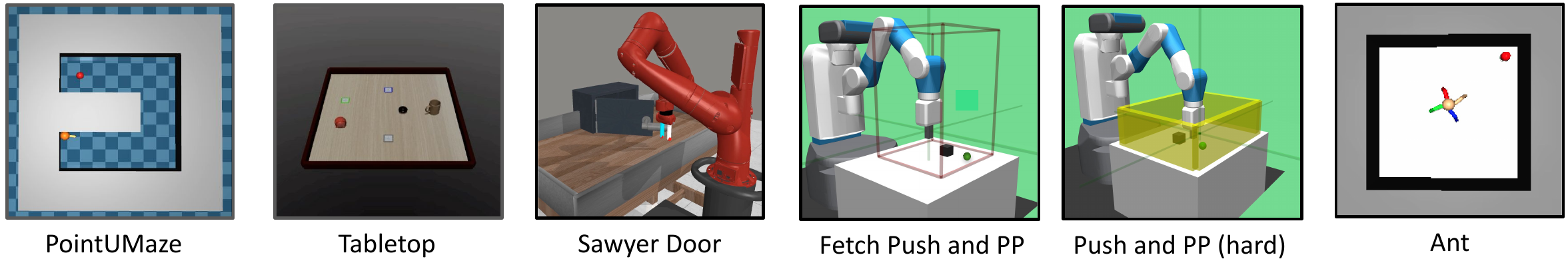}
    \caption{We evaluate \method on eight reset-free tasks ranging from navigation to manipulation. PP is short for Pick\&Place.}
    \label{fig:tasks}
\end{figure}

\textbf{Environments:} We evaluate \method and baselines on eight tasks (see \cref{fig:tasks}).
We select five tasks from IBC's evaluation suite~\citep{kim2023free} of six tasks; (PointUMaze, Tabletop, Sawyer Door, Fetch Push and PP, Fetch Reach is omitted  because it is trivially solvable). Next, we increased the complexity of the two hardest tasks from IBC, Fetch Push and Fetch Pick\&Place, by extending the size of the workspace, replacing artificial workspace limits (which cause unrealistic jittering behavior near the limits, see the website for videos) with real walls, and evaluating on harder goal states (i.e. Pick\&Place goals only in the air rather than including ones on the ground).
% To prevent the block from falling off the table in Fetch tasks, the IBC authors artificially limited the block position with block position constraints. This resulted in unrealistic jittering behavior near the limits (see the website \footnote{\url{\projectwebsite}}). We removed the artificial joint constraints and surrounded the table with physical walls, adapted two Fetch tasks to a more challenging and realistic setting, where the state space is larger and the block is harder to get unstuck from corners, i.e. Push (hard) and Pick\&Place (hard).
% In Pick\&Place (hard), the agent is only evaluated on goals that are in the air (unlike Pick\&Place that IBC uses, $50\%$ of goals is on the ground). 
In addition, we contributed a difficult locomotion task, Ant, which is adapted from the PEG codebase \citep{hu2023planning}. 

\begin{itemize}
    \item \textbf{PointUMaze}: A point-mass agent navigates a U-shape maze through continuous acceleration commands. During evaluation, the agent starts from the bottom-left corner and is tasked to reach the top-left corner. 
    \item \textbf{Tabletop Manipulation}:  The agent needs to grab and move the mug to one of the four goal locations. The initial state is always fixed and the goal state is uniformly sampled from four fixed locations. 
    \item \textbf{Sawyer Door}: The agent controls a Sawyer robot arm  to close the door in an open position. During the reset-free training, it needs to learn to close the door and open the door again to practice. The door is opened to 60 degrees for evaluation. 
    \item \textbf{Fetch Push\&PP}: The agent commands a Fetch robot arm to push / pick\&place the object initialized at the center of the table to goal locations. The environment is taken from IBC's evaluation suite, which modified the original environment from~\cite{plappert2018multi}. To prevent the block from falling off the table, the IBC authors artificially limited the block position with block position constraints.
    \item \textbf{Push\&PP(hard)}:  
    Using block position constraints (in Fetch Push\&PP) resulted in unrealistic jittering behavior near the limits. To avoid this, we removed the artificial joint constraints and surrounded the table with physical walls. Furthermore, we enable the usage of the grippers (disabled in IBC's version) to permit picking behaviors (i.e. useful for resetting), at the cost of increased action space and exploration difficulty.
    \item \textbf{Ant}: The 4-legged ant agent needs to navigate in a square room to a given goal, which is uniformly located in the top-right corner. The initial state is at the center point with randomness. It is adapted from~\cite{hu2023planning}, with changing the U-shape maze into a square room.
\end{itemize} 

Most methods are run with 5 seeds, and the mean performance and standard error are reported. During the evaluation, the performance on tasks with randomly sampled goals from $\rho_{g^*}$ is measured by averaging over 10 episodes. See \cref{supp:experimental_details} for more experimental details.

\subsection{Results}
\label{section:results}
\begin{figure*}[!htb]
    \centering
    \includegraphics[width=0.95\textwidth]{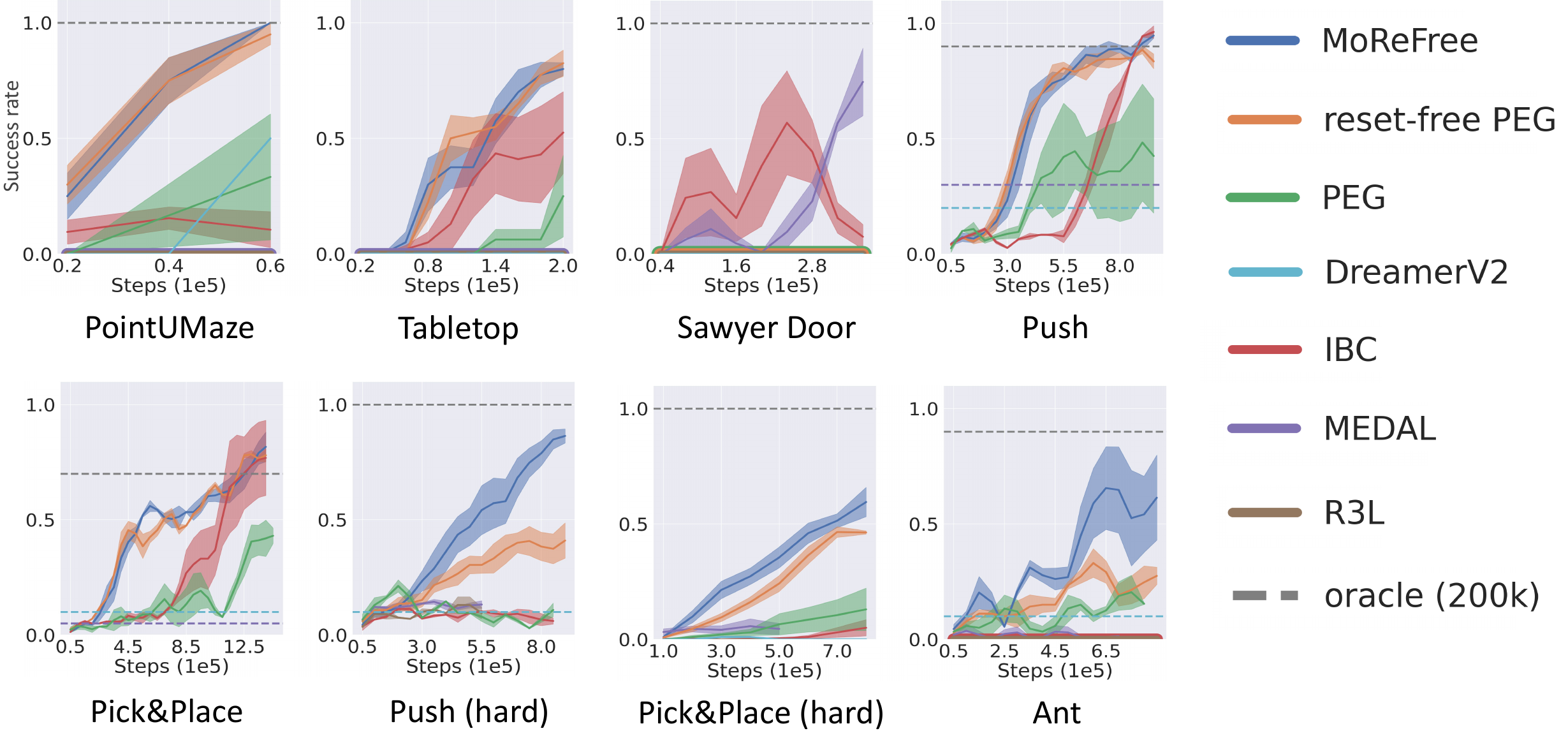}
    \caption{Two reset-free MBRL methods (\method and reset-free PEG) significantly outperform baselines in 7/8 tasks. However, directly applying MBRL methods (PEG and DreamerV2) works poorly. In 4 tasks, only MBRL methods are able to learn meaningful behavior, showcasing MBRL's sample efficiency in the reset-free setting. \method outperforms reset-free PEG in the 3 most difficult tasks.
    } 
    \label{fig:main_results}
\end{figure*}

% 1.  Model-based RL “out of the box” is a strong reset-free baseline. Simple usage of PEG already outperforms model-free RL in 6/7 (if we include IBC Fetch). MoReFree is always better than PEG. We do notice that in one environment, model-based RL methods in general struggle (see Appendix).

As shown in Fig~\ref{fig:main_results}, two reset-free model-based methods (MoReFree and reset-free PEG), without demonstrations or access to environmental reward, outperform other baselines with privileged access to supervision in both final performance and sample efficiency in 7/8 tasks.  We observe that the two reset-free MBRL methods learn good behaviors: the pointmass agent hugs the wall of the UMaze to minimize travel time and the Fetch robot deftly pushes and picks up the block into multiple target locations. \method is always competitive with or outperforms reset-free PEG, with large gains in the 3 hardest tasks: Push (hard) by $45\%$, Pick\&Place (hard) by $13\%$ and Ant (hard) by $36\%$.  We observe that \method learns non-trivial reset behaviors such as picking and pushing blocks back into the center of the table for the hard variants of the Fetch manipulation tasks. However, the original PEG performs poorly, suggesting that directly applying episodic MBRL methods in a reset-free setting without adaptations yields suboptimal results. See the website for videos of \method and baselines. 

In many tasks, the baselines fail to learn at all. We believe this is due the low sample budget, which may be too low for the baselines to fully explore the environment and learn the proper resetting behaviors necessary to train the actual task policy. In \cref{supp:ibc_fetch}, we increased the training budget by $3\times$ for the IBC baseline and it still fails, underscoring the difficulty of the tasks and the sample-efficiency gains of \method and MBRL.
% and the complexity of reset-free training where the agent must not only learn the evaluation behavior but also reset behavior to stabilize its training distribution. We noticed that in the hard Fetch tasks, even with extended training budgets, baselines fail to learn. 
% In \cref{supp:ibc_fetch}, we run baselines with extended training budgets in the hard manipulation environments and they still fail to learn. 
On the other hand, we noticed that one environment, Sawyer Door, seemed particularly hard for MBRL agents to solve. We hypothesize that the dynamics of the task are hard to model, resulting in performance degradation for model-based approaches (see \cref{supp:sawyer_door} for more analysis).

\subsection{Analysis}
% 2. MoReFree works better in more complex environments. We show that MoReFree works better in the environments with harder reset dynamics, easily getting stuck, etc. Show all the analysis here.

% 3. What components of MoReFree improve its performance? We show that both parts of MoReFree are necessary, one without the other is bad (similar to the the rebuttal response).
To explain the performance differences between \method and baselines, we closely analyze the exploration behaviors.

\begin{figure}[ht]
    \centering
    \includegraphics[width=0.95\textwidth]{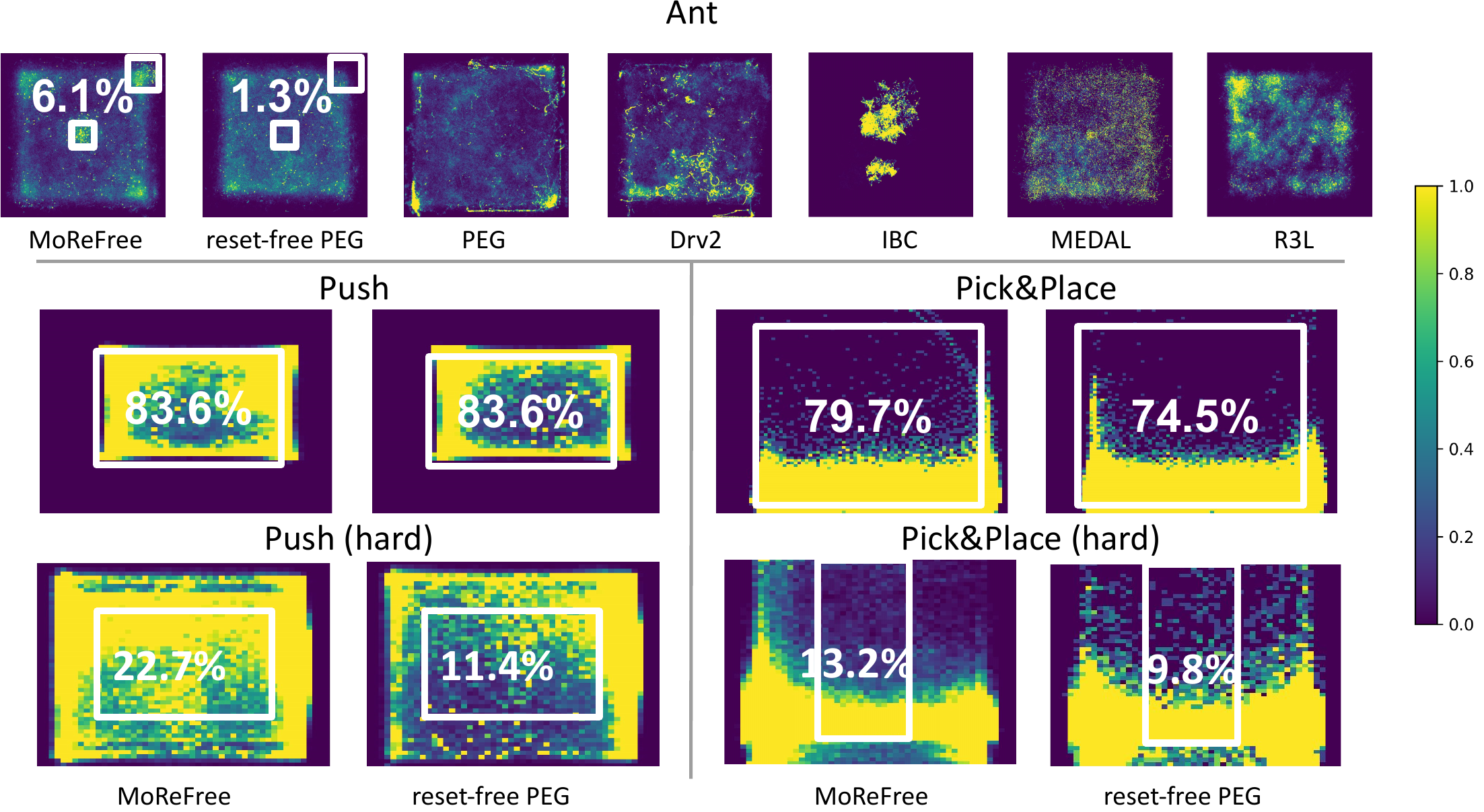}
    \caption{State visitation heatmaps of different agents. White areas are task-relevant states (including initial and goal state distributions) and we overlay the percentages of task-relevant states. reset-free MBRL methods explore more and in harder environments, \method experiences more task-relevant states.
    }
    \label{fig:main_heatmap}
\end{figure}

\textbf{\method focuses on task-relevant states.}  In \cref{fig:main_heatmap} we visualize the state visitation heatmaps of methods in various environments, and also compute the percentage of ``task-relevant'' states (initial and goal regions, highlighted with white borders). We highlight two trends. First, the heatmaps show that \method and reset-free PEG explore thoroughly while baselines have more myopic exploration patterns, as seen in the Ant heatmaps at the top.

Next, performance differences between reset-free PEG and \method are intuitively explained by the amount of task-relevant data collected by each agent. In easier environments like Push or Pick\&Place where both reset-free PEG and \method encounter similar amounts of task-relevant states, the performance is roughly similar between reset-free PEG and \method. But in harder environments (Ant, Push (hard), Pick\&Place (hard)) with larger state spaces and more complicated resetting dynamics, \method  collects $1.3 - 5\times$ more task-relevant data and has large performance gains over reset-free PEG. By experiencing more task-relevant states and training policies on them in imagination, \method policies are more suited towards succeeding at the episodic evaluation criteria. See \cref{supp:heatmaps} for additional visualizations.

% In the easier variants of Push and Pick\&Place, both PEG and \method and encounter similar amounts of task-relevant states.  In the harder environments (Ant, Push (hard), Pick\&Place (hard)) with larger state spaces and more complicated dynamics, \method collects $1.3 - 5\times$ more task-relevant data. This characterizes the performance difference between \method and PEG - by experiencing more task-relevant states and training policies on them in imagination, \method policies are more suited towards succeeding at the episodic evaluation criteria. In easier environments where PEG can collect a similar amount of task-relevant states, this performance difference lessens. See \cref{supp:heatmaps} for additional visualizations.

\begin{figure}[htb]
    \centering
    \includegraphics[width=0.95\textwidth]{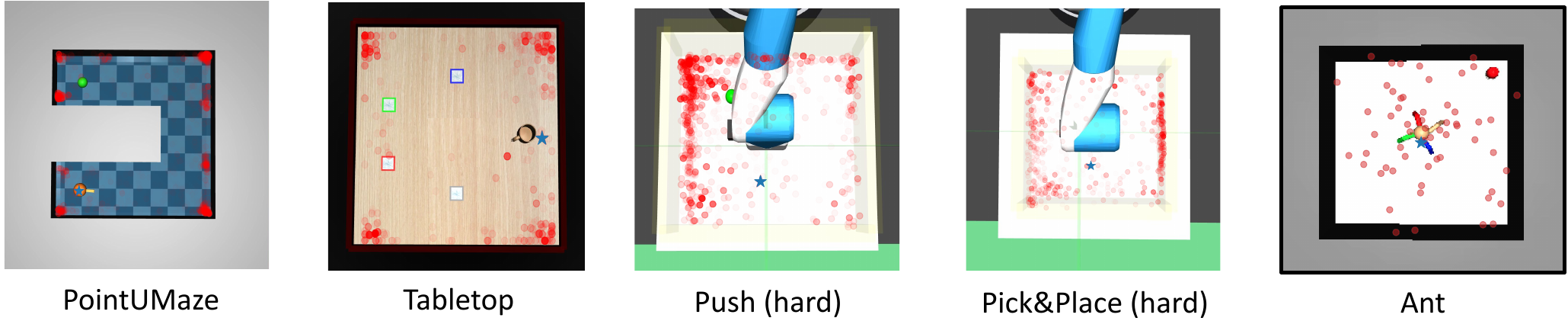}
    \caption{We visualize the start position (red dots) of  successful ``Back'' trajectories of \method's Back-and-Forth Go-Explore, where $\pi^G_\theta$ is directed to reset the environment. The color intensity of the dots correspond to state density over the last 100 steps.
    }
    \label{fig:backward}
\end{figure}

\textbf{\method effectively resets.} Next, we investigate the qualitative behavior of \method's Back-and-Forth Go-Explore. 
To see if ``Back'' trajectories help free the agent from the sink states, we analyze the replay buffer of \method for the environments, and plot the starting locations of the agent / object up to 100 timesteps before  a successful ``Back'' trajectory is executed in \cref{fig:backward}.   The color intensity of the dots correspond to state density over the last 100 steps (i.e. dark red means the agent / object has rested there for a while). We observe that the starting locations (red dots) of the agent / object are in corners or next to walls in all environments. 
This suggests that these areas act as sink states, where the agent / object would remain for long and waste time.
We observe that \method learns reset behaviors like picking the block out of corners and walls in Fetch Push and Fetch Pick\&Place. See detailed videos of the reset behavior on the website.

\subsection{Ablations}
\label{section:ablations}
\begin{wrapfigure}[17]{}{0.5\textwidth}
 %\hspace{0.5cm}
 \centering
    \includegraphics[width=77mm]{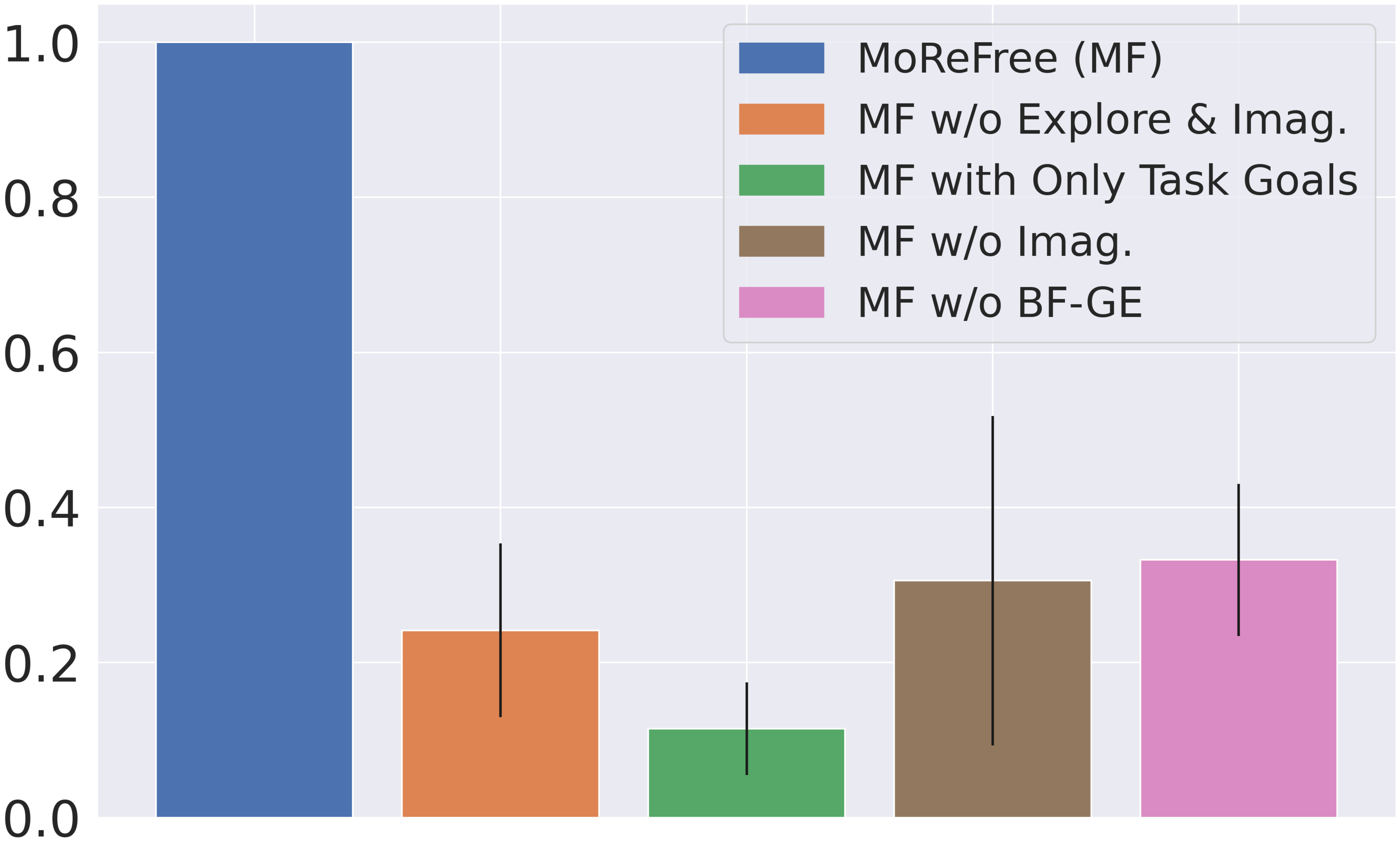}
    \caption{Ablations on 5 variants of \method over 3 hard environments, Push (hard), Pick\&Place (hard) and Ant, with normalized final performance. 
    }
    \label{fig:ablation}
\end{wrapfigure}

To justify our design choices, we ablate the two mechanisms of \method, the back-and-forth exploration and task-relevant goal-conditioned policy training, and plot the results in \cref{fig:ablation}. 

First, removing all mechanisms (\textbf{MF w/o Explore \& Imag.}) reduces to reset-free PEG, and we can see a large gap in performance.  Next, \textbf{MF with Only Task Goals} sets $\alpha=1$, which causes an extreme bias towards task-relevant states in the exploration and policy training. This also degrades performance, due to the need for strong exploration in the reset-free setting. Examinations of more values for $\alpha$ can be found in \cref{supp:hyperparameters}.

Finally, we isolate individual components of \method. First, we disable Back-and-Forth Go-Explore by disallowing the sampling of initial or evaluation goals during Go-Explore. Only exploratory goals are used in Go-Explore for this ablation (named \textbf{MF w/o BF-GE}). Next, in \textbf{MF w/o Imag.} we turn off the initial / evaluation goal sampling in imagination, so only random replay buffer goals are used to train $\pi^G_\theta$.  
We see that both variants perform poorly. This is somewhat intuitive, as the two components rely on each other. 
Having both forms a synergistic cycle where 1) the goal-conditioned policy’s optimization is more focused towards reaching initial / goal states, and 2) the exploration is biased towards reaching initial / goal states by using the goal-conditioned policy we just optimized in step 1. If we remove one without the other, then the cycle breaks down.
In \textbf{MF w/o Imag.}, Back-and-Forth Go-Explore will suffer since $\pi^G_\theta$ trained on random goals cannot reliably reach initial / evaluation goals. 
In \textbf{MF w/o BF-GE}, the exploration strategy will not seek initial / evaluation states, resulting in an inaccurate world model and degraded policy optimization. In summary, the ablations show that \method's design is sound and is the major factor behind its success in the reset-free setting. 
See \cref{supp:ablations} for details.

\section{Conclusion and Future Work}
\label{section:conclusion}
% \ehc{Other limitations to consider: unsupervised reset-free setting, integrating demos / reward functions, high dimensional images}
% \zyc{To mention limitation of our method, MoReFree doesn't work envs like sawyer door, etc.}
As a step towards reset-free training, we adapt model-based methods to the reset-free setting and demonstrate their superior performance. Specifically, we show that with minor modifications, unsupervised MBRL method substantially outperforms the state-of-the-art model-free baselines tailored for the reset-free setting while being more autonomous (requires less supervision like environmental reward or demonstrations). We then identify a limitation of unsupervised MBRL in the reset-free setting (over-exploration on task-irrelevant states), and propose \method to address such limitations by focusing model-based exploration and goal-conditioned policy training on task-relevant states. 
We conduct a thorough experimental study of \method and baselines over 8 tasks, and show considerable performance gains over the MBRL baseline and prior state-of-the-art reset-free methods. 
% Building on this, we propose \method, to address the limitations of such a direct extension, i.e. over-exploration on task-irrelevant states. By focusing model-based exploration and goal-conditioned policies on task-relevant states, \method shows enhanced performance over the MBRL backbone in more difficult tasks. 

Despite its overall success, \method is not without limitations. Being a model-based approach, it inherits all associated disadvantages. For example, we believe Sawyer Door is a task where learning the dynamics is harder than learning the policy (see \cref{supp:sawyer_door}), disadvantaging MBRL approaches. Next, \method uses a fixed percentage of task-relevant goals for exploration and imagination, whereas future work could consider an adaptive curriculum. Finally, scaling \method to high-dimensional observations and real-world applications would be natural extensions.
We hope \method inspires future efforts in increasing autonomy in RL.

\bibliography{main}
\bibliographystyle{tmlr}

%%%%%%%%%%%%%%%%%%%%%%%%%%%%%%%%%%%%%%%%%%%%%%%%%%%%%%%%%%%%%%%%%%%%%%%%%%%%%%%
%%%%%%%%%%%%%%%%%%%%%%%%%%%%%%%%%%%%%%%%%%%%%%%%%%%%%%%%%%%%%%%%%%%%%%%%%%%%%%%
% APPENDIX
%%%%%%%%%%%%%%%%%%%%%%%%%%%%%%%%%%%%%%%%%%%%%%%%%%%%%%%%%%%%%%%%%%%%%%%%%%%%%%%
%%%%%%%%%%%%%%%%%%%%%%%%%%%%%%%%%%%%%%%%%%%%%%%%%%%%%%%%%%%%%%%%%%%%%%%%%%%%%%%
\newpage
\appendix
\section{Broader Impacts}
\label{supp:impacts}
As we increase the autonomy of RL agents, the possibility of them acting in unexpected ways to maximize reward increases. The unsupervised exploration coupled alongside the learned reward functions further add to the unpredictability; neither mechanisms are very interpretable.
As such, we expect research into value alignment, interpretability, and safety to be paramount as autonomy in RL improves.

\section{Experimental Details}
\label{supp:experimental_details}

\subsection{Environments}
\label{supp:envs}
\textbf{PointUMaze:} The state space is 7D and the action space is 2D. The initial state is $(0, 0)$, which located in the bottom-left corner, and noise sampled from $\mathcal{U}(-0.1, 0.1)$ is added when reset. The goal during the evaluation is always located in at the top-left corner of the U-shape maze. The maximum steps during the evaluation is 100. Hard reset will happen after every $2e5$ steps. In the whole training process we performed, it only reset once at the beginning of the training. Taken from the IBC \citep{kim2023free} paper.

\textbf{Tabletop:} The state space is 6D, and the action space is 3D. During the evaluation, four goal locations are sampled in turn, the initial state of the agent is always fixed and located in the center of the table. The maximum steps during the evaluation is 200. Hard reset will happens after every $2e5$ steps. In the whole training process we performed, it only reset once at the beginning of the training. Taken from the EARL \citep{sharma2021autonomous} benchmark and also used in the IBC paper.

\textbf{Sawyer Door:} The state space is 7D and the action space is 4D. The position of door is initialized to open state ($60$ degree with noise sampled from $(0, 18)$ degree) and the goal is always to close the door (0 degree). The arm is initialized to a fixed location. Maximum number of steps is 300 for the evaluation. Hard reset will happen after every $2e5$ steps. In the whole training process we performed, it resets twice. Taken from the EARL \citep{sharma2021autonomous} benchmark and also used in the IBC paper.

\textbf{Fetch Push and Pick\&Place:} The state space is 25D and action space is 4D. These are taken from the IBC paper. Authors converted the original Fetch environments to a reversible setting by defining a constraint on the block position. The initial and goal distributions are identical to the original Fetch Push and Pick\&Place. More details can be found in the IBC paper.

\textbf{Push (hard)}: Different from the original Fetch Push task, in our case walls are added to prevent the block from dropping out of the table. The workspace of the robot arm is also limited. The block is always initialized to a fixed location, and goal distribution during the evaluation is $\mathcal{U}(-0.15, 15)$. Fetch Push used in the IBC paper, the block is limited by joint constraint, which shows unrealistic jittering behaviors near the limits (we observe such phenomenon by running model-based go-explore, the exploration policy prefers to always interact with the block and keep pushing it towards the limit boundary, see videos on our project website~\footnote{\url{\projectwebsite}}). Meanwhile, the gripper is blocked, which makes the task easier. In our case, we release the gripper and it can now open and close again which add two more dimension of the state space. We found it is important to release the gripper in our version of Push task, when the block is in corners, it will need to operate the gripper to drag the block escape from corners. The maximum steps the agent can take in 50 during the evaluation. Hard reset will happen after every $1e5$ steps. In the whole training process we performed, it resets 5 times in total. 

\textbf{Pick\&Place (hard)}: We add walls in the same way as we did for Push (hard). We make it more difficult by only evaluating the agent on goals that are in the air. Then it has to learn to perform picking behavior properly, whereas goals on the ground can just be solved by pushing. The goal will be uniformly sampled from a $5\times5\times10$ cm cubic area above the table. It has the same observation space, action space, initial state and maximum steps with Fetch Push described above. Hard reset will happens after every $1e5$ steps. In the whole training process we performed, it resets 5 times in total. See the visual difference between our Pick\&Place and IBC's in ~\cref{fig:tasks}. Since the workspace of the robot is limited within the walls as well in Push (hard) and Pick\&Place (hard), when the block gets stuck in corners, the robot needs to precisely move to the corner and bring the block back. In contrast, the robot in IBC's version can move to everywhere, being able to create various circumstance to solve such difficult position.

\textbf{Ant:} We adapt the AntMaze task from environments\footnote{\url{https://github.com/edwhu/mrl}} codebase of PEG and change the shape of the maze to square, also change the evaluation goal distribution to be a uniform distribution $\mathcal{U}(2,3)$ for both x and y location, which lies on the top-left corner of the square. The ant is always initialized to the center point (0, 0) of the square to start from, with uniform noise ($\mathcal{U}(-0.1, 0.1)$) added. The state space is 29D and the action space is 8D. The maximum steps for evaluation is 500. Hard reset will happen after every $2e5$ steps. In the whole training process we performed, it reset 4 times in total.

\subsection{Baseline Implementations}
\label{supp:baselines}
\textbf{PEG}: We use the official implementation of PEG\footnote{\url{https://github.com/penn-pal-lab/peg}} and only optimize the exploratory goal distribution once at the beginning of each reset-free training episode, i.e. $H_G$ and $H_E$ are set to half of the reset-free episode length.

\textbf{reset-free PEG:} We extend the official implementation of PEG\footnote{\url{https://github.com/penn-pal-lab/peg}} to reset-free setting by 1) set $H_G$ and $H_E$ to half of the evaluation episode length; 2) optimizing the goal distribution every $H_G+H_E$ steps; 3) keeping all other hyperparameters the same as MoReFree.

\textbf{IBC:} We use the official implementation from authors\footnote{\url{https://github.com/snu-larr/ibc_official}} and keep hyperparameters unchanged.

\textbf{DreamerV2}: We use the official implementation of PEG. In order to reduce it to DreamerV2~\citep{hafner2020dreamerv2}, we remove the exploration policy and only execute goal-conditioned policy for the whole reset-free episode. During imagination training, the goal-conditioned policy is only trained on the evaluation goal distribution.

\textbf{MEDAL:} We follow the official implementation of MEDAL\footnote{\url{https://github.com/architsharma97/medal}} and use the deafult setting for experiments. Since MEDAL requires demonstrations, for tasks from EARL benchmark, demonstrations are provided. For other environments, we generate demonstrations by executing the final trained \method to collect data. 30 episodes are generated for each task.

\textbf{R3L:} We implement R3L agent by modifying the FBRL agent from MEDAL codebase. The backward policy is replaced by an exploration policy trained using the random network distillation (RND) objective \citep{burda2018exploration}. The RND implementation
we follow is from DI-engine\footnote{\url{https://opendilab.github.io/DI-engine/12_policies/rnd.html}}.

\textbf{Oracle:} This is a episodic SAC agent, we use the implementation from MEDAL codebase and keep all the hyper-parameters unchanged.

\textbf{\method:} Our agent is built on the model-based go-explore method PEG~\citep{hu2023planning}, we extend their codebase by adding back-and-forth goal sampling procedure and training on evaluation initial and goal states in imagination goal-conditioned policy training. See our codebase in the supplemental.

\subsection{Hyperparameters}
\label{supp:hyperparameters}
Train ratio (i.e. Update to Data ratio) is an important hyper-parameter in MBRL. It controls how frequently the agent is trained. Every $n$ steps, a batch of data is sampled from the replay buffer, the world model is trained on the batch, and then policies and value functions are trained in imagination. In all our experiments, we only vary $n$ on different tasks. See the table below for different values on different tasks we used through experiments. \method also introduces a new parameter $\alpha$, which we keep $\alpha=0.2$ for all tasks and did not tune it at all. All other hyperparameters we keep the same as the original code base.
\begin{table}[!htb]
    \caption{Different train ratio we used for different tasks. We keep all other hyperparameters the same as default ones.}
    \centering
    \begin{tabular}{|c|c|c|c|c|c|c|c|}
    \hline
     PointUMaze    & 2 & Sawyer Door   & 5 & Tabletop      & 1 & Fetch Push    & 2 \\
     \hline
     Fetch Pick\&Place  & 2 & Push (hard)    & 2 & Pick\&Place (hard) & 2 & Ant    & 2 \\
     \hline
    \end{tabular}
    % \label{tab:my_label}
\end{table}

\begin{wrapfigure}[15]{}{0.3\textwidth}
\vspace{-0.8cm}
\hspace{-0.5cm}
 \centering
    \includegraphics[width=40mm]{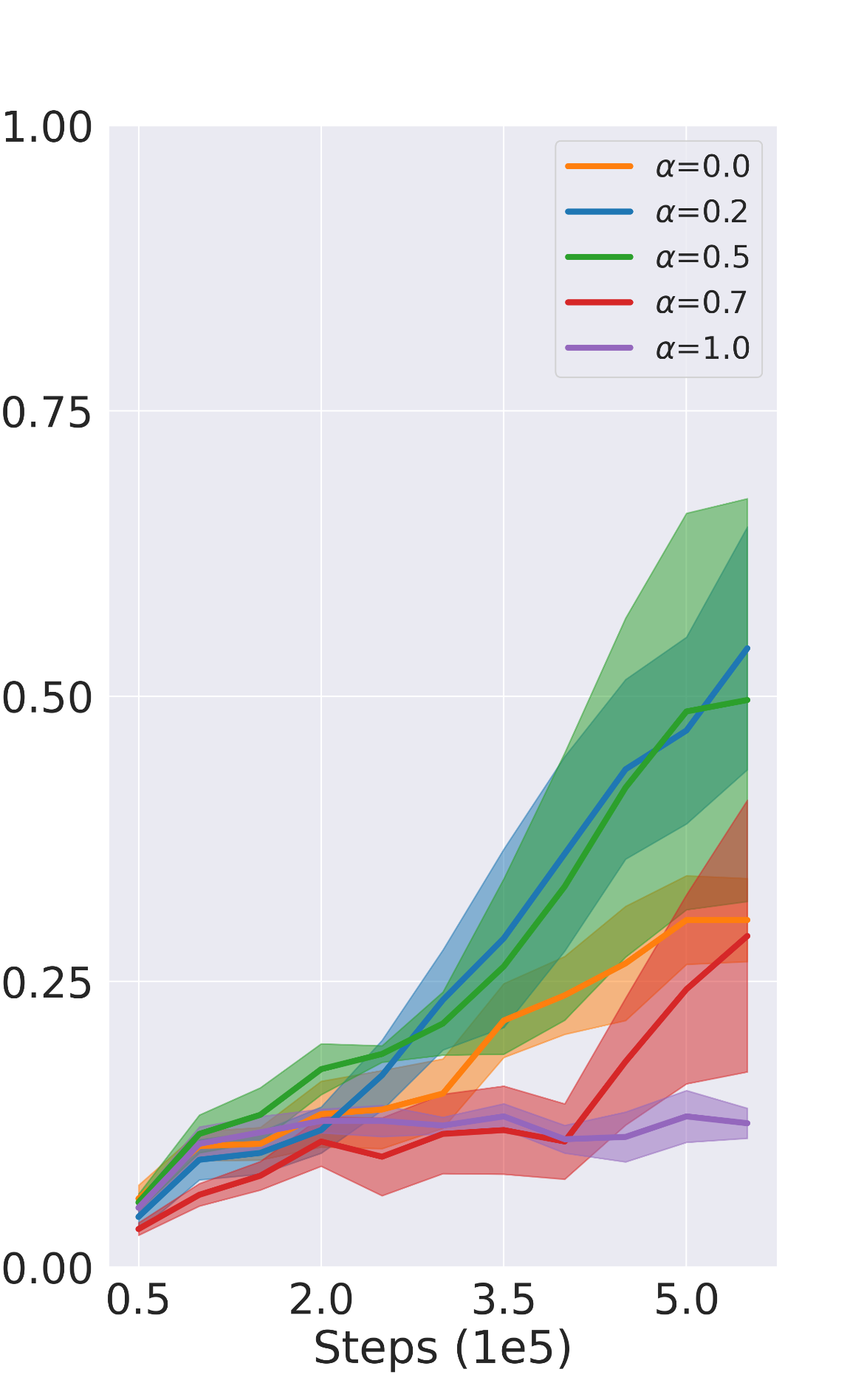}
    \caption{Performance of \method with different values of $\alpha$ in Push (hard).}
    \label{fig:alpha}
\end{wrapfigure}

\textbf{Different values for $\alpha$}. We examine different values of $\alpha$ in \method on Fetch Push task, which affects how much \method focuses on task-relevant goals in exploration and imagination. In \cref{fig:alpha}, we see that introducing a moderate amount of task-relevant goals ($\alpha$=0.2, $\alpha$=0.5) results in sensible performance, while too many task-relevant goals ($\alpha$=0.7, $\alpha$=1.0) degrades performance. We use the same value of alpha, 0.2, across all tasks, which showcases \method’s consistency.

\subsection{Results Clarification}
\label{supp:results_clarification}
In Push and Pick\&Place results, we retrieved the final performance of MEDAL directly from the IBC paper (dashed purple lines) and did not have time to run R3L in these two environments. R3L is shown to be a lot worse than MEDAL in the MEDAL paper and performs obviously bad in other tasks shown in \cref{fig:main_results}. In Push (hard) and Pick\&Place (hard), we ran R3L and MEDAL with less budget since other methods clearly outperform and their learning curves do not show any evidence for going up.

\subsection{Resource Usage}
%  GPU, CPU, training time, etc.
\label{supp:resource}
We submit jobs on a cluster with Nvidia 2080, 3090 and A100 GPUs. Our model-based experiments take 1-2 days to finish, and the model-free baselines take half day to one day to run. 

\section{Method Details}
\label{supp:method}
Here, we provide a more in-depth exposition of self-supervised goal-reaching reward function~\citep{hartikainen2019dynamical} that is used for training the goal-conditioned policy, PEG~\citep{hu2023planning} that is used for generating exploratory goal distribution, and P2E~\citep{sekar2020planning} that is used for training the exploration policy. 

\subsection{Self-supervised Goal-reaching Reward Function}
\label{supp:reward fn}
\method does not require environment reward functions, instead it learns a distance function $d_w$ for training the goal-conditioned policy. $d_w$ is trained by sampling pairs of states $s_t, s_{t+k}$ from an imagined rollout of the goal-conditioned policy and predicting the distance $k/H$, where $H$ is the maximum distance equal to the imagination horizon. Then the reward is defined as $r(s_t, g)=-d_w(s_t,g)$. 

\subsection{Exploratory Goal Distribution} 
\label{supp:peg}
We use PEG to generate the exploratory goal distribution $\rho_E$ in \method. PEG generates goals that have high exploration potentials. To evaluate a goal $g$, the goal-conditioned policy is rolled out for $K$ trajectories $\tau_k$ within the learned world model. Then the terminal state exploration value for each trajectory with the learned exploration value function $V_{\theta}^E(s_T^k)$ is estimated, where $s_T^k$ is the last state of the trajectory $\tau_k$. Then the estimates are averaged. 

The goal variable $g$ is optimized using model predictive path integral control (MPPI). First, $N$ goal candidates $g$ are sampled from an initial distribution. These candidates are then evaluated as described above. This averaged exploration value acts as the ``score" for the goal candidate. Once we have scores for each goal candidate, a Gaussian distribution is fit according to the rule:
\begin{equation}
    \mu_t=\frac{\sum_{k=0}^N(e^{\gamma\cdot V_k})(g_k)}{\sum_{k=0}^N(e^{\gamma \cdot V_k})}
\end{equation}

where $\gamma$ is the reward weight hyperparameter. We then sample candidates from the computed Gaussian, and repeat the process for multiple iterations. After the last iteration, $\rho_E$ is defined as the computed Gaussian.

\subsection{Plan2Explore}
\label{supp:p2e}
The world model $\hat{\mathcal{T}}_{\theta}$ consists of the following components:
\begin{equation}
\begin{split}
        &\text{encoder: } e_t=\text{enc}_\theta(x_t)
        \hspace{1.6cm} \text{posterior: } q_\theta(s_t|s_{t-1}, a_{t-1},e_t) \\
        &\text{dynamics: } p_\theta(s_t|s_{t-1},a_{t-1}) 
        \hspace{0.9cm}
        \text{decoder: } p_\theta(x_t|s_t) \\
    \end{split}
\end{equation}
The model states $s_t$ contain a deterministic component $h_t$ and a stochastic component $z_t$. 

P2E is the objective we used to train the exploration policy, and it encourages the agent to visit states that can improve the world model the most. We train an ensemble of 1-step models to predict the next model state from the current model state:
\begin{equation}
    \text{Ensemble}:\hspace{2cm} f(s_t, \theta^k)=\hat{z}_{t+1}^k \hspace{2cm} \text{for } k=1...K
\end{equation}

Then the exploration reward is the variance of the ensemble predictions averaged across dimension of the model state, $r(s_t)=\frac{1}{N} \sum_n \text{Var}_{\{k\}}[f(s_t, \theta_k)]_n$.

\color{black}
\section{More Visualizations on Replay Buffer}
\label{supp:heatmaps}
\begin{figure}
    \centering
    \includegraphics[scale=0.3]{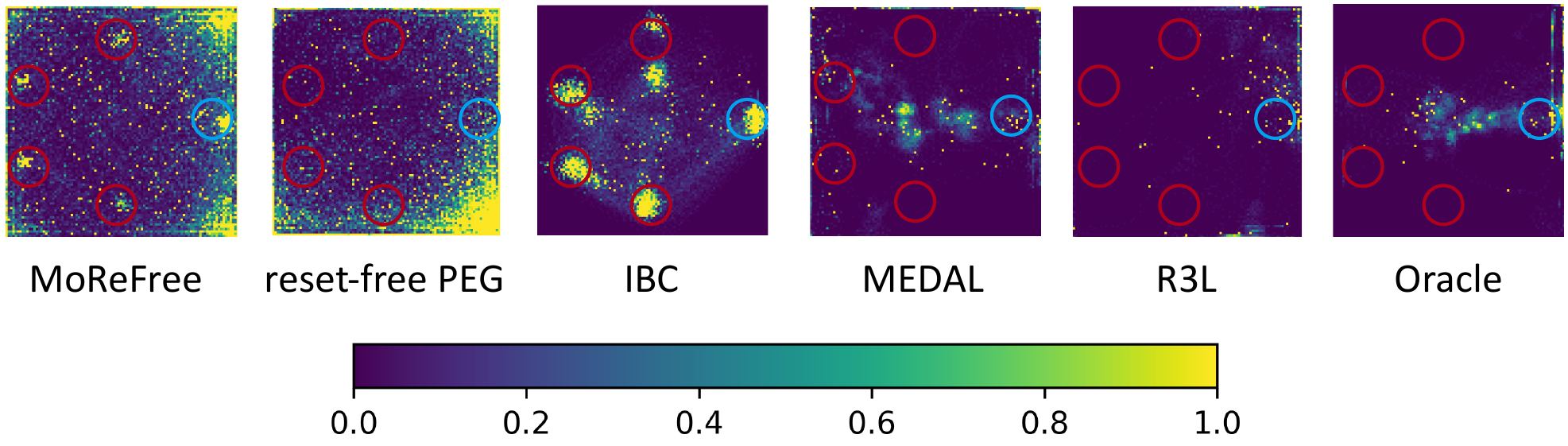}
    \caption{XY state visitation heatmap of the mug in Tabletop of various approaches. MoReFree’s heatmap shows high state diversity while retaining high visitation counts near the task-relevant states (red circles are goal states, the blue circle is the initial state). reset-free PEG also shows diverse exploration, but it over-explores the bottom-right corner which is entirely task-irrelevant. IBC's bi-directional curriculum leads the exploration shuttles between the initial state and goal states, but fails to explore well. All other methods fail to explore, visited states mostly cluster in few spots.}
    \label{fig:tabletop_heatmap}
\end{figure}

\begin{figure}
    \centering
    \includegraphics[scale=0.5]{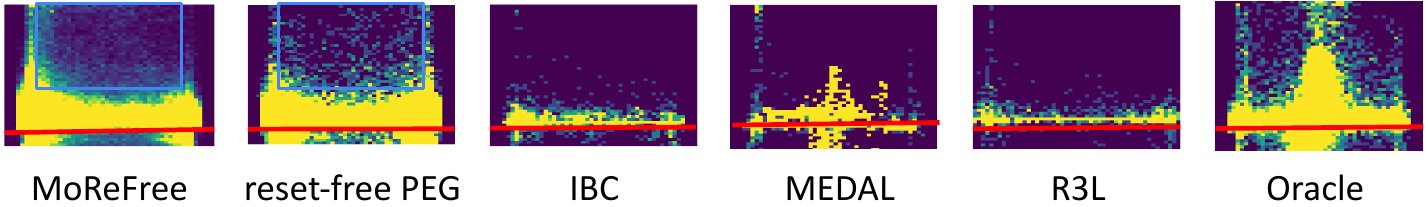}
    \caption{XZ state visitation heatmap of the block in Pick\&Place (hard). States above the red line are in the air, which are crucial for solving the picking task. Two MBRL methods collect more data diversely in the air, while other reset-free methods barely pick up the block.}
    \label{fig:xz_heatmap}
\end{figure}

\begin{figure}[!htb]
    \centering
    \includegraphics[scale=0.25]{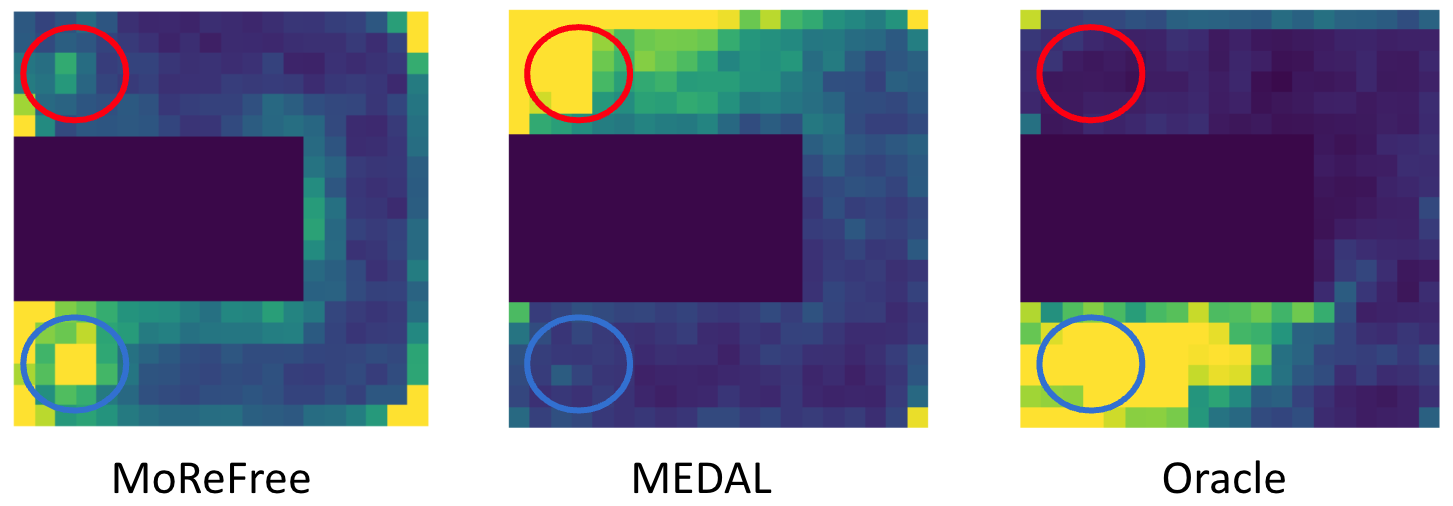}
    \caption{State visitation heatmap on point maze. \method has special focuses on both initial state (blue circles) corner and goal state (red circles), while explore much uniformly. MEDAL collects lots of data near the goal state and little data on the initial state. Both MEDAL and Oracle explore less extensively.}
    \label{fig:point_heatmap_dif_agents}
\end{figure}

\begin{figure}[!htb]
    \centering
    \includegraphics[scale=0.22]{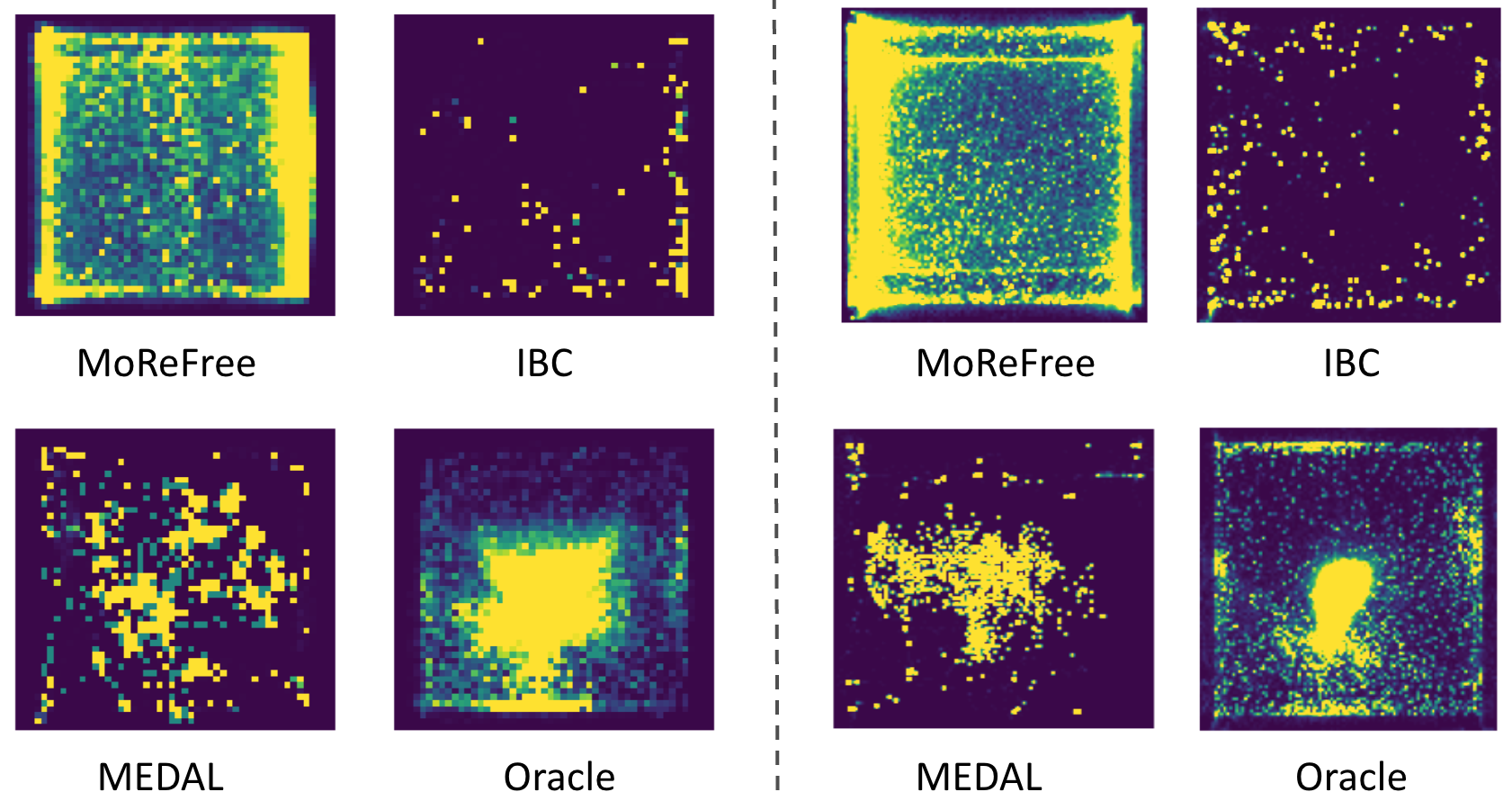}
    \caption{Block state visitation heatmap on Fetch Push (left) and Fetch Pick\&Place (right) of different agents. \method better explores the whole state space, while IBC and MEDAL do not have too much interactions with the block, thus lighted areas are scattered everywhere.}
    \label{fig:pp_heatmap}
\end{figure}

We visualize the replay buffer of different agents on more tasks. See \cref{fig:tabletop_heatmap} for XY location of the mug in Tabletop, \cref{fig:point_heatmap_dif_agents} for XY location data of the agent in PointUMaze, \cref{fig:xz_heatmap} for XZ location of the block in Pick\&Place (hard) and \cref{fig:pp_heatmap} for XY location data of the block in Push (hard) and Pick\&Place (hard). Overall, we see \method explores the whole state space better. Meanwhile, due to back-and-forth procedure, \method collects more data near initial / goal states, which are important for the evaluation. However, IBC, MEDAL, R3L and Oracle all fail to explore well; their heatmaps are mostly populated with low visitation cells.

\section{Detailed Ablations}
\begin{figure}
    \centering
    \includegraphics[scale=0.2]{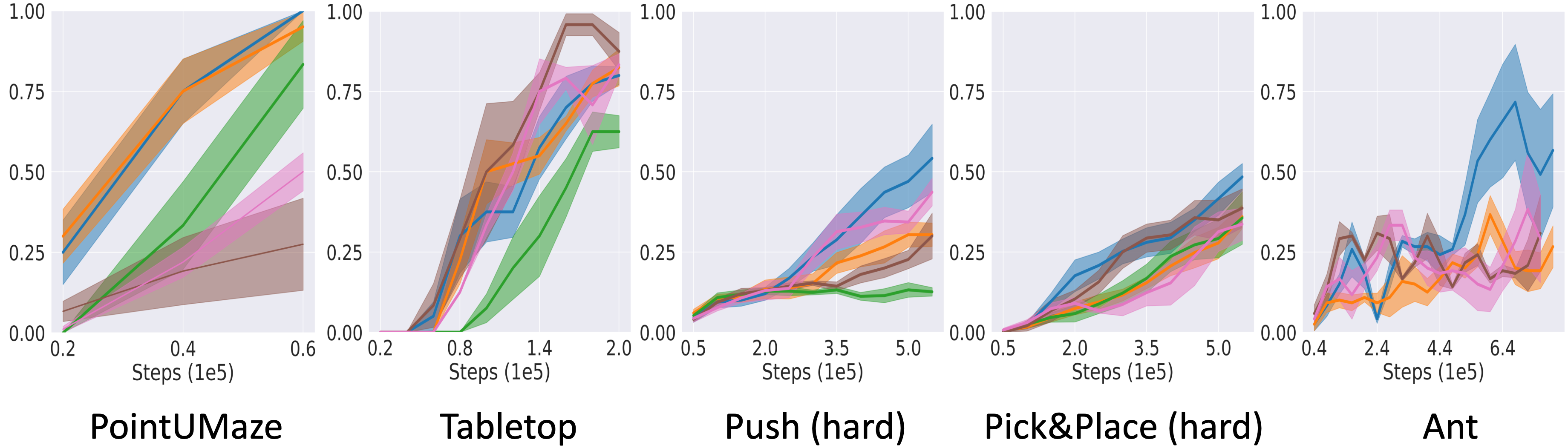}
    \caption{Learning curves of ablation study on 5 tasks. We see different components contribute differently in different tasks. For instance, in Tabletop, \textbf{MF w/o Imag.} even performs better than \method, maybe because the whole state space can be explored quickly, then randomly sampling states from the replay buffer as goals for training already has good coverage on evaluation initial / goal states.}
    \label{fig:all_ablations}
\end{figure}

We report learning curves for each variant agent we ablate in \cref{section:ablations} on every task in ~\cref{fig:all_ablations}. Since \method does not learn at all in Saywer Door task, we exclude the ablation for it. In each task, \method is better or on par with all other ablations. Through learning curves, we see different components contribute differently on different tasks.

\begin{figure}[!htb]
    \centering
    \includegraphics[scale=0.2]{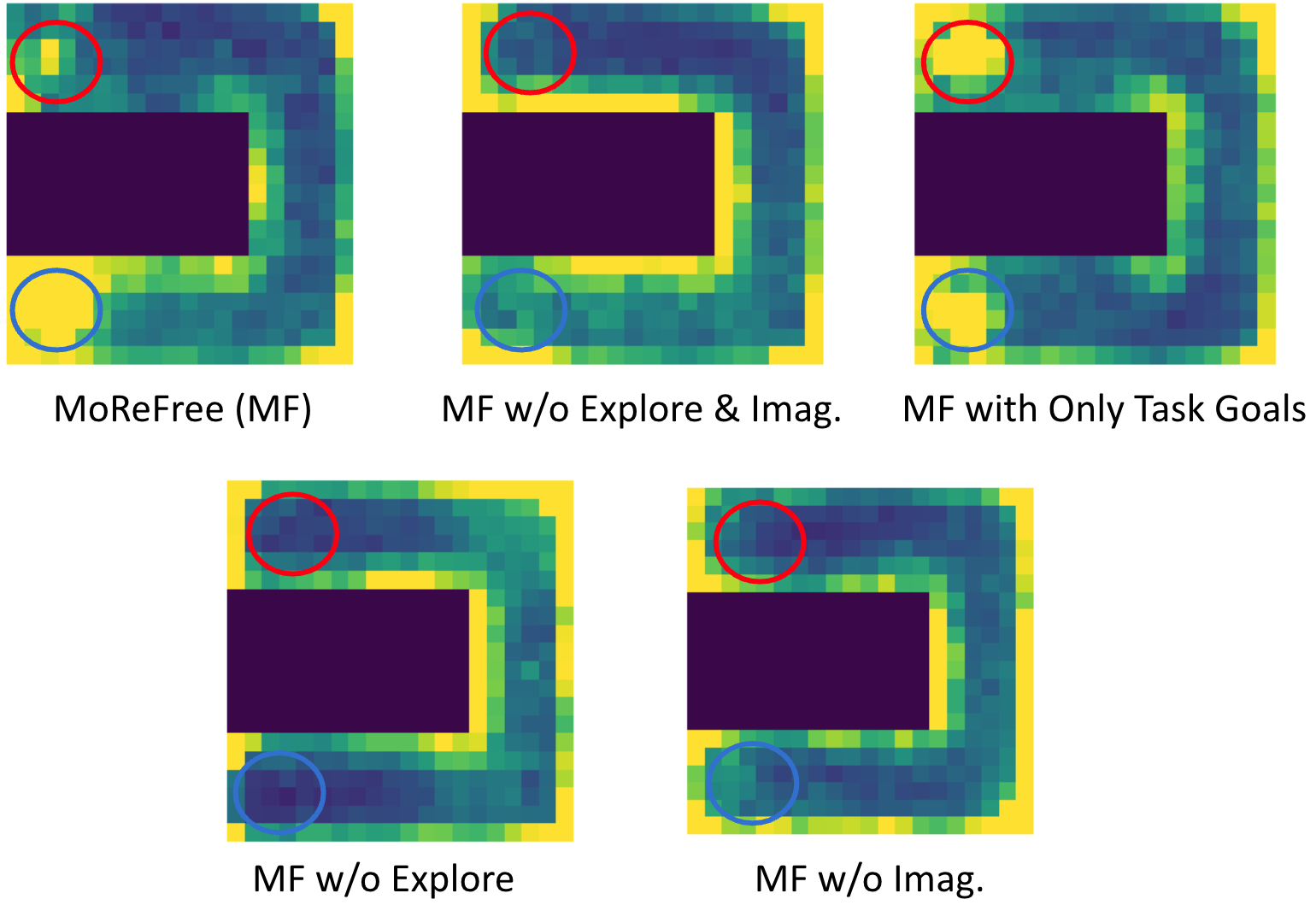}
    \caption{State visitation heatmap on PointUMaze task of all ablations. Red circles are evaluation goal states and blues are initial states. We see \method collect good amount of data near initial / goal states while stronger exploration. MF w/o Explore and MF w/o Imag. could not gather task-relative data, which further causes poor performance.}
    \label{fig:point_heatmap}
\end{figure}

We further analyze the ablation on PointUMaze as an example by visualizing the replay buffer of different variants, see ~\cref{fig:point_heatmap}. In the performance on PointUMaze from ~\cref{fig:all_ablations}, sampling exploratory goals for data collection is important (MF w/o Explore \& Imag. outperforms other ablations). But we see in ~\ref{fig:point_heatmap}, MF w/o Explore \& Imag. does not have focus on the initial / goal state which we care about for the evaluation, which makes it slightly worse than \method. MF with Only Task Goals has a strong preference on initial / goal state, we think it is because in the later phase of the training when the agent is able to solve the task, it goes back-and-forth consistently to collect data. But in the early phase of the training, it might lack exploration which causes the degraded performance compare with \method. MF w/o Explore and MF w/o Imag. only either go to initial / goal state for data collection and do not practice on it during the imagination training, or practice without really going, which both does not form the positive cycle, and end up with poor performance.  
\label{supp:ablations}

\section{MBRL on Sawyer Door}
\label{supp:sawyer_door}
We investigate why two MBRL methods fail on Sawyer Door tasks. Note that \method is able to solve intermediate goals such as closing the door in some angles, but is unable to solve the original IBC evaluation goal (see website for more videos).

\begin{figure}[!htb]
    \centering
    \includegraphics[height=5cm]{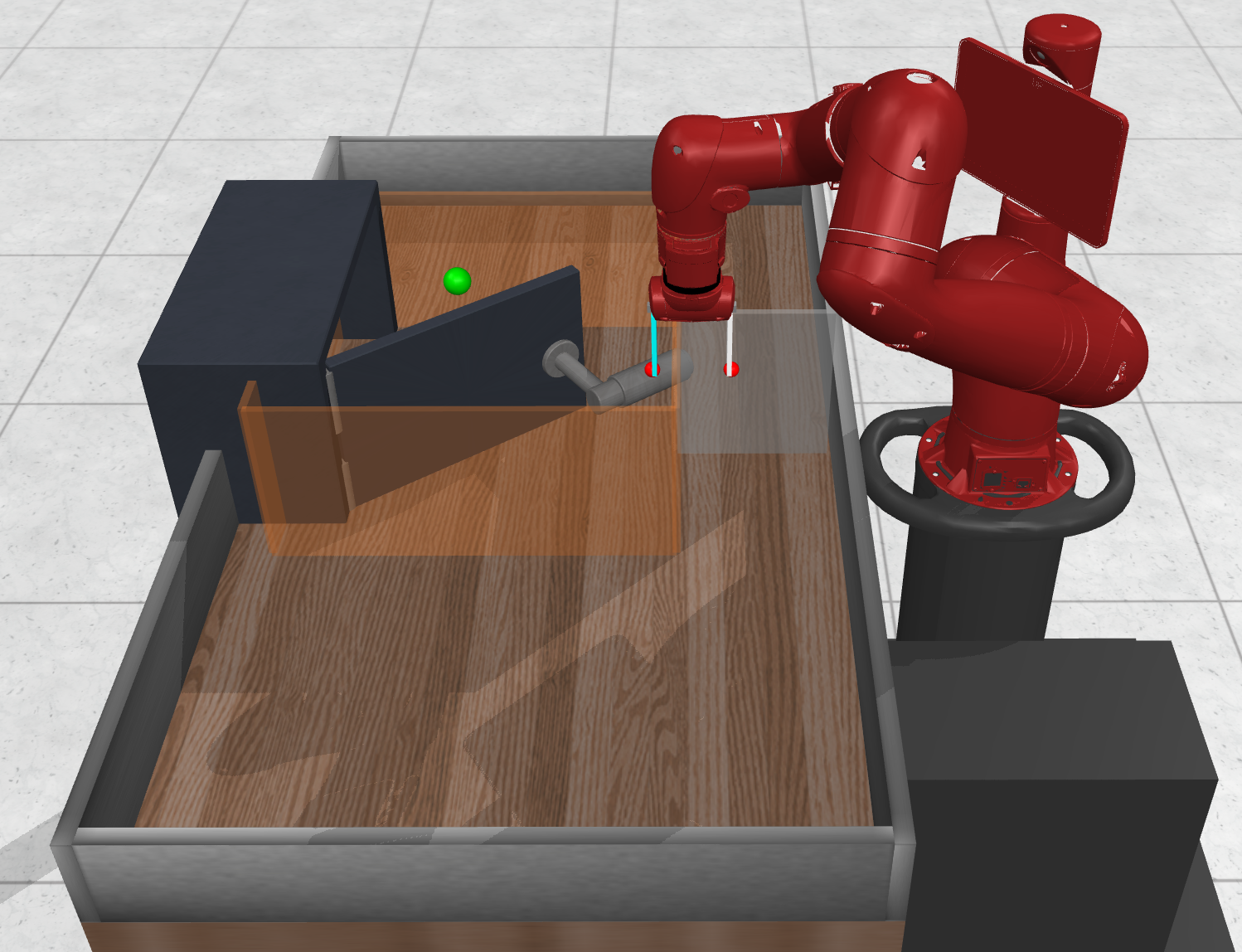}
    \caption{Simplified version of Sawyer Door. Orange walls show the limited workspace for the robot arm, and a grey wall is added to limit the movement of the door. The door can only move to maximum $60$ degrees.}
    \label{fig:simple_door}
\end{figure}

We simplify Sawyer Door task by limiting the movement range of the robot to a box and also having a block holds the door to prevent it from opening it too much, see~\cref{fig:simple_door}. Although MBRL methods are trained on the simplified environment, we see learning curves on Sawyer Door are completely flat in ~\cref{fig:main_results}, compared with other baselines trained on the original task. We wonder why MBRL methods can show the same performance and gain benefits as it does in other environments.

\method and reset-free PEG use DreamerV2 as backbone agents and extend it to reset-free settings. We hypothesize that Dreamer itself, even under the episodic setting with task reward function, would not work well. If that's the case, then MBRL methods in the reset-free setting with self-supervised reward function would almost certainly not work either. For example, if the backbone agent cannot model the dynamics precisely, then policy learning, dynamical distance reward learning, will be degraded.

\begin{figure}[h]
    \centering
    \includegraphics[height=4cm]{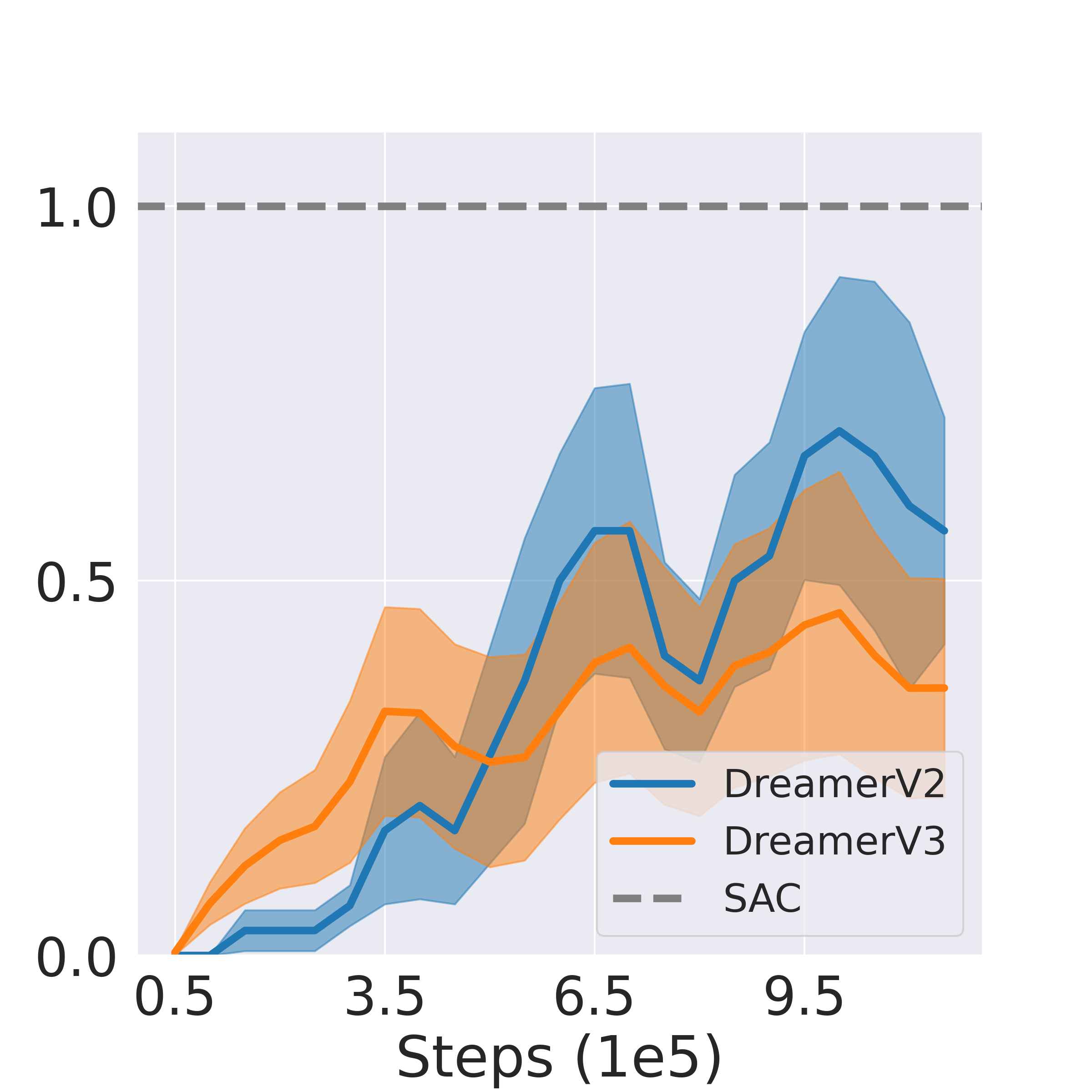}
    \caption{Performance of DreamerV2 and V3 on episodic Sawyer Door task. SAC can solve the task in 200k steps, while after 1 million steps MBRL is still not able to steadily solve the task.}
    \label{fig:dreamer}
\end{figure}
We then run the underlying MBRL backbones under the episodic setting. \cref{fig:dreamer} shows DreamerV2~\footnote{\url{https://github.com/danijar/dreamerv2}}, and  Dreamerv3~\footnote{\url{https://github.com/danijar/dreamerv3}}  struggle to solve the task, while model-free method SAC can steadily solve the task after 200k steps. 
This might be a potential reason that MBRL methods do not work on the more difficult reset-free setting. 
We hypothesize that the combination of the sparse environmental reward and dynamics of the door result in a hard prediction problem for world modelling approaches. We leave further investigation for the future work.

\section{More Analysis on Fetch Environments}
\label{supp:ibc_fetch}
\begin{figure}
    \centering
    \includegraphics[scale=0.15]{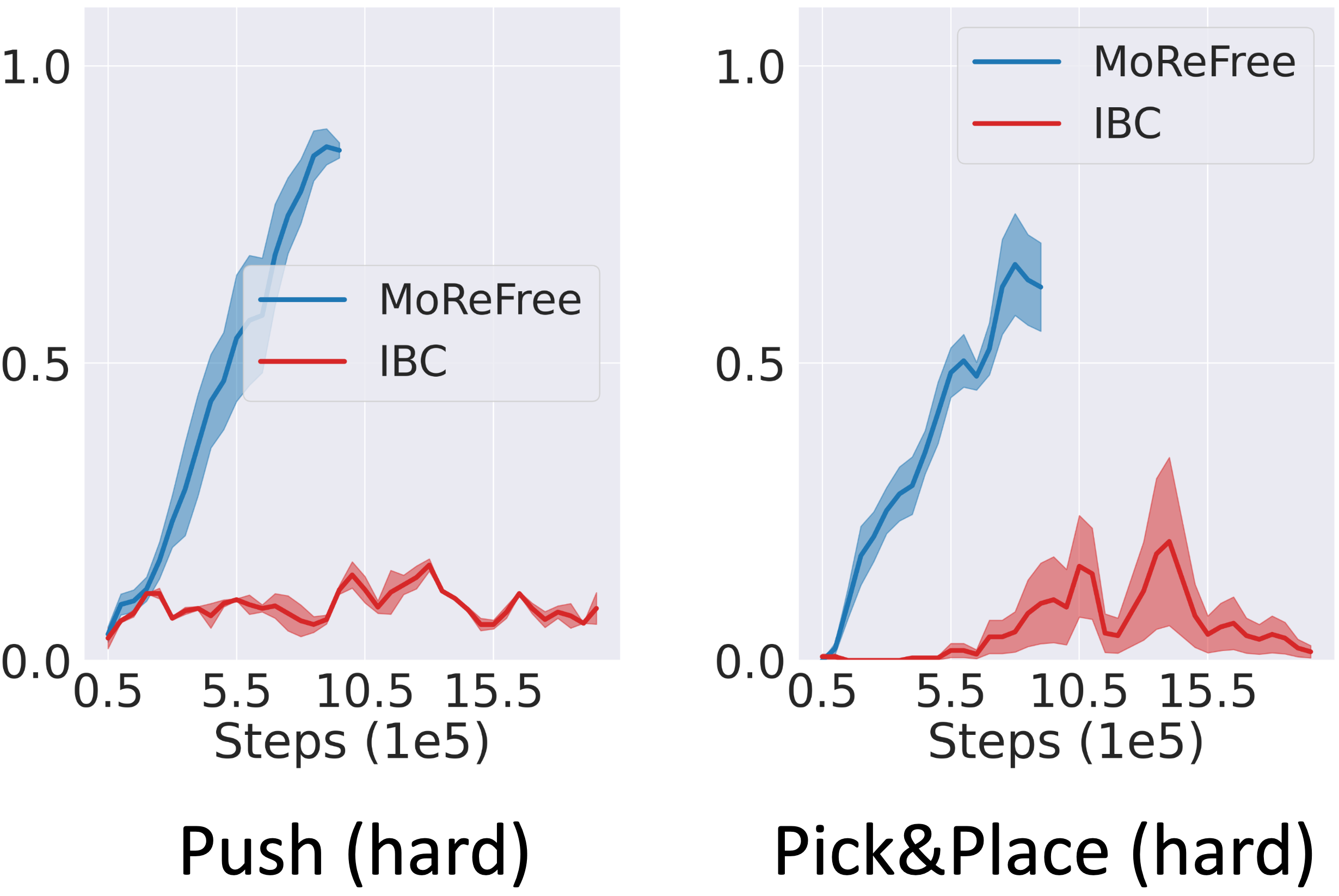}
    \caption{Longer training of IBC in our Fetch tasks, where the state space is larger and artificial constraints are replaced with surrounded walls. IBC still can not learn meaningful behaviors.}
    \label{fig:more_fetch}
\end{figure}

Although IBC gains good final performance in Push and Pick\&Place, it starts learning late compared with MBRL methods and fails entirely in our harder versions. We suspect IBC might need more computational budget to start learning in harder tasks. Thus we train IBC with two millions environment steps and results in \cref{fig:more_fetch} show that it still fails to solve the harder version of Push.

\begin{figure}
    \centering
    \includegraphics[scale=0.15]{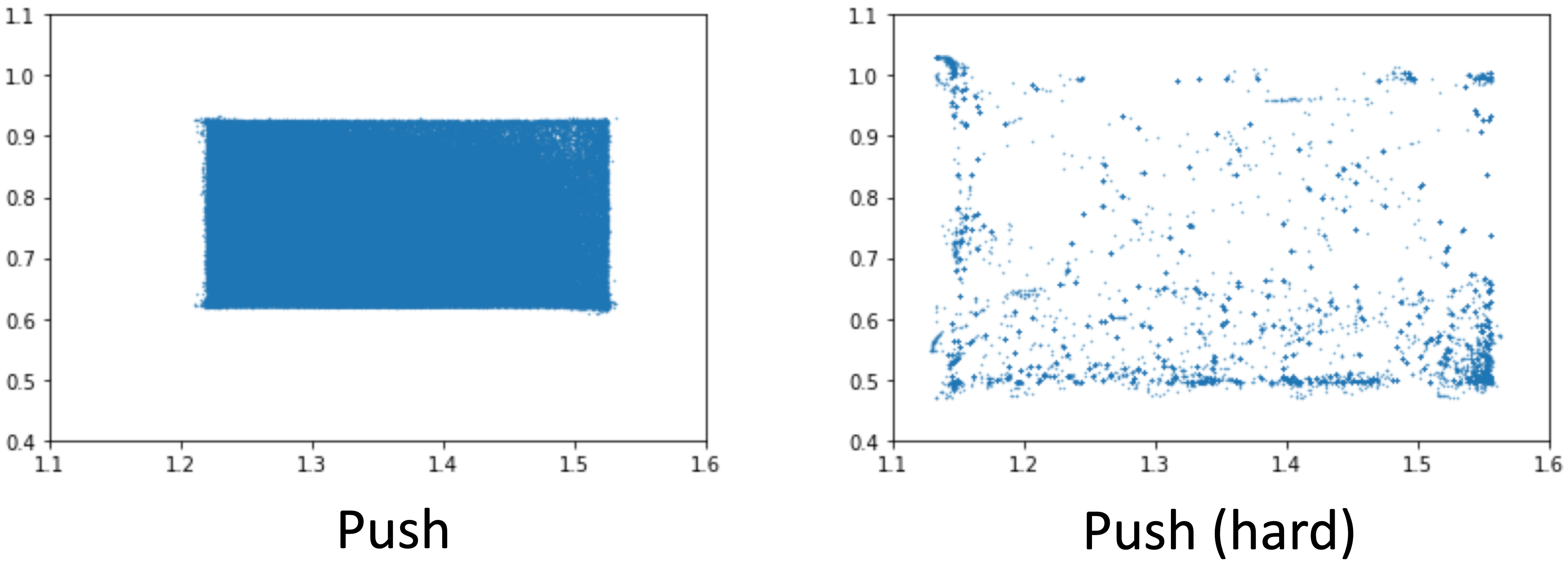}
    \caption{XY location of the block collected by IBC on Push (hard) and its original version (Push). IBC covers the whole state space very well in Push while fails in Push (hard), where the block stays for long time in corners or areas next to walls.}
    \label{fig:rb_push_ibc}
\end{figure}
\cref{fig:rb_push_ibc} shows 600k data of the obejct (XY view) collected by IBC on our Push (hard) and IBC's Push. We see the block stays in corners or next to walls a lot in Push (hard), while goes everywhere and covers the whole space in IBC's Push, indicating object interaction is more difficult in Push (hard) due to the larger state space, surrounded walls and limited work space. In IBC's Push, the block can bounce back when it hits the limit of joint constraints. However, in Push (hard), the block needs to be explicitly brought back from the corner or walls, requiring more sophisticated behaviors. Meanwhile, larger size of the limited area (our version is $3\times$ larger than IBC's.) also increases the difficulty of the task.

\section{Analysis on R3L}
\label{supp:r3l}
\begin{figure}[!htb]
    \centering
    \includegraphics[scale=0.3]{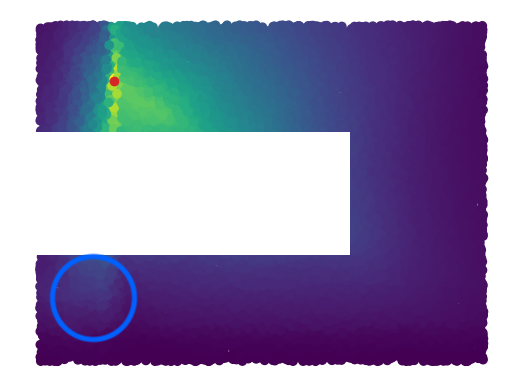}
    \caption{Outputs of the learned classifier on the whole state space. Due to the smoothness of the output scope, states near the initial state (blue circle) also have higher values.}
    \label{fig:r3l}
\end{figure}

R3L trains two policies, one for reaching the goal and another that brings the agent to novel states. The goal-reaching policy is trained using a learned classifier to classify the goal state and other states. Original R3L takes images as inputs, thus the trained classifier can successfully classify goal images from random state images. In our work, we use low-dimensional state input. Outputs of the trained classifier on the whole state space of PointUMaze is shown in \cref{fig:r3l}. We see that the classifier learns to output higher values for states close to the goal state (red dot) and lower values for states further away. Nonetheless, due to the smoothness of the output scope, states near the initial state (blue circle) that are numerically closer but spatially further to the goal state also have higher values. R3L agent trained using such reward function will always tend to follow states with higher values to the corner instead of going forward. See the website for more videos. These trained reward functions are misleading for learning reasonable policies which result in poor performance we see in \cref{fig:main_results}.
\end{document}

%% file: math_commands.tex
\usepackage{amsmath,amsfonts,bm}

\def\eqref#1{equation~\ref{#1}}
\def\1{\bm{1}}

\DeclareMathAlphabet{\mathsfit}{\encodingdefault}{\sfdefault}{m}{sl}
\SetMathAlphabet{\mathsfit}{bold}{\encodingdefault}{\sfdefault}{bx}{n}

\newcommand{\E}{\mathbb{E}}

%% file: main.bbl
\begin{thebibliography}{36}
\providecommand{\natexlab}[1]{#1}
\providecommand{\url}[1]{\texttt{#1}}
\expandafter\ifx\csname urlstyle\endcsname\relax
  \providecommand{\doi}[1]{doi: #1}\else
  \providecommand{\doi}{doi: \begingroup \urlstyle{rm}\Url}\fi

\bibitem[Baranes \& Oudeyer(2013)Baranes and Oudeyer]{baranes2013active}
Adrien Baranes and Pierre-Yves Oudeyer.
\newblock Active learning of inverse models with intrinsically motivated goal exploration in robots.
\newblock \emph{Robotics and Autonomous Systems}, 61\penalty0 (1):\penalty0 49--73, 2013.
\newblock ISSN 0921-8890.
\newblock \doi{https://doi.org/10.1016/j.robot.2012.05.008}.
\newblock URL \url{https://www.sciencedirect.com/science/article/pii/S0921889012000644}.

\bibitem[Burda et~al.(2018)Burda, Edwards, Storkey, and Klimov]{burda2018exploration}
Yuri Burda, Harrison Edwards, Amos Storkey, and Oleg Klimov.
\newblock Exploration by random network distillation.
\newblock \emph{arXiv preprint arXiv:1810.12894}, 2018.

\bibitem[Ecoffet et~al.(2021)Ecoffet, Huizinga, Lehman, Stanley, and Clune]{ecoffet2021first}
Adrien Ecoffet, Joost Huizinga, Joel Lehman, Kenneth~O Stanley, and Jeff Clune.
\newblock First return, then explore.
\newblock \emph{Nature}, 590\penalty0 (7847):\penalty0 580--586, 2021.

\bibitem[Eysenbach et~al.(2017)Eysenbach, Gu, Ibarz, and Levine]{eysenbach2017leave}
Benjamin Eysenbach, Shixiang Gu, Julian Ibarz, and Sergey Levine.
\newblock Leave no trace: Learning to reset for safe and autonomous reinforcement learning.
\newblock \emph{arXiv preprint arXiv:1711.06782}, 2017.

\bibitem[Eysenbach et~al.(2021)Eysenbach, Salakhutdinov, and Levine]{eysenbach2021clearning}
Benjamin Eysenbach, Ruslan Salakhutdinov, and Sergey Levine.
\newblock C-learning: Learning to achieve goals via recursive classification.
\newblock In \emph{International Conference on Learning Representations}, 2021.
\newblock URL \url{https://openreview.net/forum?id=tc5qisoB-C}.

\bibitem[Florensa et~al.(2018)Florensa, Held, Geng, and Abbeel]{florensa2018automatic}
Carlos Florensa, David Held, Xinyang Geng, and Pieter Abbeel.
\newblock Automatic goal generation for reinforcement learning agents.
\newblock In \emph{International conference on machine learning}, pp.\  1515--1528. PMLR, 2018.

\bibitem[Fu et~al.(2018)Fu, Singh, Ghosh, Yang, and Levine]{fu2018variational}
Justin Fu, Avi Singh, Dibya Ghosh, Larry Yang, and Sergey Levine.
\newblock Variational inverse control with events: A general framework for data-driven reward definition.
\newblock \emph{Advances in neural information processing systems}, 31, 2018.

\bibitem[Gupta et~al.(2021)Gupta, Yu, Zhao, Kumar, Rovinsky, Xu, Devlin, and Levine]{gupta2021reset}
Abhishek Gupta, Justin Yu, Tony~Z Zhao, Vikash Kumar, Aaron Rovinsky, Kelvin Xu, Thomas Devlin, and Sergey Levine.
\newblock Reset-free reinforcement learning via multi-task learning: Learning dexterous manipulation behaviors without human intervention.
\newblock In \emph{2021 IEEE International Conference on Robotics and Automation (ICRA)}, pp.\  6664--6671. IEEE, 2021.

\bibitem[Haarnoja et~al.(2018)Haarnoja, Zhou, Hartikainen, Tucker, Ha, Tan, Kumar, Zhu, Gupta, Abbeel, et~al.]{haarnoja2018soft}
Tuomas Haarnoja, Aurick Zhou, Kristian Hartikainen, George Tucker, Sehoon Ha, Jie Tan, Vikash Kumar, Henry Zhu, Abhishek Gupta, Pieter Abbeel, et~al.
\newblock Soft actor-critic algorithms and applications.
\newblock \emph{arXiv preprint arXiv:1812.05905}, 2018.

\bibitem[Hafner et~al.(2019)Hafner, Lillicrap, Ba, and Norouzi]{hafner2019dreamer}
Danijar Hafner, Timothy Lillicrap, Jimmy Ba, and Mohammad Norouzi.
\newblock Dream to control: Learning behaviors by latent imagination.
\newblock \emph{arXiv preprint arXiv:1912.01603}, 2019.

\bibitem[Hafner et~al.(2020)Hafner, Lillicrap, Norouzi, and Ba]{hafner2020dreamerv2}
Danijar Hafner, Timothy Lillicrap, Mohammad Norouzi, and Jimmy Ba.
\newblock Mastering atari with discrete world models.
\newblock \emph{arXiv preprint arXiv:2010.02193}, 2020.

\bibitem[Haldar et~al.(2023)Haldar, Pari, Rai, and Pinto]{haldar2023teach}
Siddhant Haldar, Jyothish Pari, Anant Rai, and Lerrel Pinto.
\newblock Teach a robot to fish: Versatile imitation from one minute of demonstrations.
\newblock \emph{arXiv preprint arXiv:2303.01497}, 2023.

\bibitem[Hartikainen et~al.(2019)Hartikainen, Geng, Haarnoja, and Levine]{hartikainen2019dynamical}
Kristian Hartikainen, Xinyang Geng, Tuomas Haarnoja, and Sergey Levine.
\newblock Dynamical distance learning for semi-supervised and unsupervised skill discovery.
\newblock \emph{arXiv preprint arXiv:1907.08225}, 2019.

\bibitem[Hu et~al.(2023)Hu, Chang, Rybkin, and Jayaraman]{hu2023planning}
Edward~S. Hu, Richard Chang, Oleh Rybkin, and Dinesh Jayaraman.
\newblock Planning goals for exploration.
\newblock In \emph{The Eleventh International Conference on Learning Representations}, 2023.
\newblock URL \url{https://openreview.net/forum?id=6qeBuZSo7Pr}.

\bibitem[Kim et~al.(2022)Kim, hyeon Park, Cho, and Kim]{kim2022automating}
Jigang Kim, J~hyeon Park, Daesol Cho, and H~Jin Kim.
\newblock Automating reinforcement learning with example-based resets.
\newblock \emph{IEEE Robotics and Automation Letters}, 7\penalty0 (3):\penalty0 6606--6613, 2022.

\bibitem[Kim et~al.(2023)Kim, Cho, and Kim]{kim2023free}
Jigang Kim, Daesol Cho, and H~Jin Kim.
\newblock Demonstration-free autonomous reinforcement learning via implicit and bidirectional curriculum.
\newblock In \emph{International Conference on Machine Learning}. PMLR, 2023.

\bibitem[Levine et~al.(2016)Levine, Finn, Darrell, and Abbeel]{levine2016end}
Sergey Levine, Chelsea Finn, Trevor Darrell, and Pieter Abbeel.
\newblock End-to-end training of deep visuomotor policies.
\newblock \emph{The Journal of Machine Learning Research}, 17\penalty0 (1):\penalty0 1334--1373, 2016.

\bibitem[Lu et~al.(2020{\natexlab{a}})Lu, Grover, Abbeel, and Mordatch]{lu2020reset}
Kevin Lu, Aditya Grover, Pieter Abbeel, and Igor Mordatch.
\newblock Reset-free lifelong learning with skill-space planning.
\newblock \emph{arXiv preprint arXiv:2012.03548}, 2020{\natexlab{a}}.

\bibitem[Lu et~al.(2020{\natexlab{b}})Lu, Mordatch, and Abbeel]{lu2020adaptive}
Kevin Lu, Igor Mordatch, and Pieter Abbeel.
\newblock Adaptive online planning for continual lifelong learning, 2020{\natexlab{b}}.
\newblock URL \url{https://openreview.net/forum?id=HkgFDgSYPH}.

\bibitem[Mendonca et~al.(2021)Mendonca, Rybkin, Daniilidis, Hafner, and Pathak]{lexa2021}
Russell Mendonca, Oleh Rybkin, Kostas Daniilidis, Danijar Hafner, and Deepak Pathak.
\newblock Discovering and achieving goals via world models, 2021.

\bibitem[Nagabandi et~al.(2020)Nagabandi, Konolige, Levine, and Kumar]{nagabandi2020deep}
Anusha Nagabandi, Kurt Konolige, Sergey Levine, and Vikash Kumar.
\newblock Deep dynamics models for learning dexterous manipulation.
\newblock In \emph{Conference on Robot Learning}, pp.\  1101--1112. PMLR, 2020.

\bibitem[Pitis et~al.(2020)Pitis, Chan, Zhao, Stadie, and Ba]{pitis2020maximum}
Silviu Pitis, Harris Chan, Stephen Zhao, Bradly Stadie, and Jimmy Ba.
\newblock Maximum entropy gain exploration for long horizon multi-goal reinforcement learning.
\newblock In \emph{International Conference on Machine Learning}, pp.\  7750--7761. PMLR, 2020.

\bibitem[Plappert et~al.(2018)Plappert, Andrychowicz, Ray, McGrew, Baker, Powell, Schneider, Tobin, Chociej, Welinder, et~al.]{plappert2018multi}
Matthias Plappert, Marcin Andrychowicz, Alex Ray, Bob McGrew, Bowen Baker, Glenn Powell, Jonas Schneider, Josh Tobin, Maciek Chociej, Peter Welinder, et~al.
\newblock Multi-goal reinforcement learning: Challenging robotics environments and request for research.
\newblock \emph{arXiv preprint arXiv:1802.09464}, 2018.

\bibitem[Pong et~al.(2019)Pong, Dalal, Lin, Nair, Bahl, and Levine]{pong2019skew}
Vitchyr~H Pong, Murtaza Dalal, Steven Lin, Ashvin Nair, Shikhar Bahl, and Sergey Levine.
\newblock Skew-fit: State-covering self-supervised reinforcement learning.
\newblock \emph{arXiv preprint arXiv:1903.03698}, 2019.

\bibitem[Sekar et~al.(2020)Sekar, Rybkin, Daniilidis, Abbeel, Hafner, and Pathak]{sekar2020planning}
Ramanan Sekar, Oleh Rybkin, Kostas Daniilidis, Pieter Abbeel, Danijar Hafner, and Deepak Pathak.
\newblock Planning to explore via self-supervised world models.
\newblock In \emph{ICML}, 2020.

\bibitem[Sharma et~al.(2021{\natexlab{a}})Sharma, Gupta, Levine, Hausman, and Finn]{sharma2021vaprl}
Archit Sharma, Abhishek Gupta, Sergey Levine, Karol Hausman, and Chelsea Finn.
\newblock Autonomous reinforcement learning via subgoal curricula.
\newblock \emph{Advances in Neural Information Processing Systems}, 34:\penalty0 18474--18486, 2021{\natexlab{a}}.

\bibitem[Sharma et~al.(2021{\natexlab{b}})Sharma, Xu, Sardana, Gupta, Hausman, Levine, and Finn]{sharma2021autonomous}
Archit Sharma, Kelvin Xu, Nikhil Sardana, Abhishek Gupta, Karol Hausman, Sergey Levine, and Chelsea Finn.
\newblock Autonomous reinforcement learning: Formalism and benchmarking.
\newblock \emph{arXiv preprint arXiv:2112.09605}, 2021{\natexlab{b}}.

\bibitem[Sharma et~al.(2022)Sharma, Ahmad, and Finn]{sharma2022state}
Archit Sharma, Rehaan Ahmad, and Chelsea Finn.
\newblock A state-distribution matching approach to non-episodic reinforcement learning.
\newblock \emph{arXiv preprint arXiv:2205.05212}, 2022.

\bibitem[Sharma et~al.(2023)Sharma, Ahmed, Ahmad, and Finn]{sharma2023self}
Archit Sharma, Ahmed~M Ahmed, Rehaan Ahmad, and Chelsea Finn.
\newblock Self-improving robots: End-to-end autonomous visuomotor reinforcement learning.
\newblock \emph{arXiv preprint arXiv:2303.01488}, 2023.

\bibitem[Smith et~al.(2019)Smith, Dhawan, Zhang, Abbeel, and Levine]{smith2019avid}
Laura Smith, Nikita Dhawan, Marvin Zhang, Pieter Abbeel, and Sergey Levine.
\newblock Avid: Learning multi-stage tasks via pixel-level translation of human videos.
\newblock \emph{arXiv preprint arXiv:1912.04443}, 2019.

\bibitem[Veeriah et~al.(2018)Veeriah, Oh, and Singh]{veeriah2018many}
Vivek Veeriah, Junhyuk Oh, and Satinder Singh.
\newblock Many-goals reinforcement learning.
\newblock \emph{ArXiv}, abs/1806.09605, 2018.

\bibitem[Xu et~al.(2020)Xu, Verma, Finn, and Levine]{xu2020continual}
Kelvin Xu, Siddharth Verma, Chelsea Finn, and Sergey Levine.
\newblock Continual learning of control primitives: Skill discovery via reset-games.
\newblock \emph{Advances in Neural Information Processing Systems}, 33:\penalty0 4999--5010, 2020.

\bibitem[Yahya et~al.(2017)Yahya, Li, Kalakrishnan, Chebotar, and Levine]{yahya2017collective}
Ali Yahya, Adrian Li, Mrinal Kalakrishnan, Yevgen Chebotar, and Sergey Levine.
\newblock Collective robot reinforcement learning with distributed asynchronous guided policy search.
\newblock In \emph{2017 IEEE/RSJ International Conference on Intelligent Robots and Systems (IROS)}, pp.\  79--86. IEEE, 2017.

\bibitem[Zhang et~al.(2020)Zhang, Abbeel, and Pinto]{zhang2020automatic}
Yunzhi Zhang, Pieter Abbeel, and Lerrel Pinto.
\newblock Automatic curriculum learning through value disagreement.
\newblock \emph{Advances in Neural Information Processing Systems}, 33:\penalty0 7648--7659, 2020.

\bibitem[Zhu et~al.(2019)Zhu, Gupta, Rajeswaran, Levine, and Kumar]{zhu2019dexterous}
Henry Zhu, Abhishek Gupta, Aravind Rajeswaran, Sergey Levine, and Vikash Kumar.
\newblock Dexterous manipulation with deep reinforcement learning: Efficient, general, and low-cost.
\newblock In \emph{2019 International Conference on Robotics and Automation (ICRA)}, pp.\  3651--3657. IEEE, 2019.

\bibitem[Zhu et~al.(2020)Zhu, Yu, Gupta, Shah, Hartikainen, Singh, Kumar, and Levine]{zhu2020ingredients}
Henry Zhu, Justin Yu, Abhishek Gupta, Dhruv Shah, Kristian Hartikainen, Avi Singh, Vikash Kumar, and Sergey Levine.
\newblock The ingredients of real-world robotic reinforcement learning.
\newblock \emph{arXiv preprint arXiv:2004.12570}, 2020.

\end{thebibliography}
